\documentclass[10pt,twocolumn,letterpaper]{article}

\usepackage[pagenumbers]{cvpr} 

\usepackage{microtype}

\renewcommand{\paragraph}[1]{\vspace{.5em}\noindent\textbf{#1}}

\DeclareMathOperator*{\argmin}{arg\,min}

\usepackage[most]{tcolorbox}
\usepackage{multicol, multirow, graphicx}
\usepackage{amsmath}
\usepackage{subcaption}
\definecolor{tabfirst}{rgb}{1, 0.7, 0.7}
\definecolor{tabsecond}{rgb}{1, 0.85, 0.7}
\definecolor{tabthird}{rgb}{1, 1, 0.7}
\usepackage[table]{xcolor}
\usepackage{tablefootnote}
\usepackage{tikz}
\usetikzlibrary{positioning}
\usepackage{amsthm}
\newcommand\blfootnote[1]{%
  \begingroup
  \renewcommand\thefootnote{}\footnote{#1}%
  \addtocounter{footnote}{-1}%
  \endgroup
}
\usepackage{array, makecell}
\newcommand{\tightcell}[2]{#1 \tiny$\pm$ #2}

\newtheorem{remark}{Remark}

\definecolor{cvprblue}{rgb}{0.21,0.49,0.74}
\usepackage[pagebackref,breaklinks,colorlinks,allcolors=cvprblue]{hyperref}

\def\paperID{7372} 
\def\confName{CVPR}
\def\confYear{2026}

\title{Training-free Mixed-Resolution Latent Upsampling\\for Spatially Accelerated Diffusion Transformers}

\author{Wongi Jeong$^{*1}$ \qquad Kyungryeol Lee$^{*1}$ \qquad Hoigi Seo$^{1}$ \qquad Se Young Chun$^{1,2\dag}$ \\
$^1$Dept. of Electrical and Computer Engineering, $^2$INMC \&  IPAI \\
Seoul National University, Republic of Korea \\
{\tt\small \{wg7139, kr.lee, seohoiki3215, sychun\}@snu.ac.kr} \\
\href{https://ignoww.github.io/RALU_project}{\textcolor{cyan}{\texttt{https://ignoww.github.io/RALU\_project}}}
\vspace{-0.5em}
}

\begin{document}
\maketitle

\begin{abstract}
    Diffusion transformers (DiTs) offer excellent scalability for high-fidelity generation, but their computational overhead poses a great challenge for practical deployment. Existing acceleration methods primarily exploit the temporal dimension, whereas spatial acceleration remains underexplored. In this work, we investigate spatial acceleration for DiTs via latent upsampling. We found that na\"ive latent upsampling for spatial acceleration introduces artifacts, primarily due to aliasing in high-frequency edge regions and mismatching from noise-timestep discrepancies. Then, based on these findings and analyses, we propose a training-free spatial acceleration framework, dubbed Region-Adaptive Latent Upsampling (RALU), to mitigate those artifacts while achieving spatial acceleration of DiTs by our mixed-resolution latent upsampling. RALU achieves artifact-free, efficient acceleration with early upsampling only on artifact-prone edge regions and noise-timestep matching for different latent resolutions, leading to up to 7.0$\times$ speedup on FLUX-1.dev and 3.0$\times$ on Stable Diffusion 3 with negligible quality degradation. Furthermore, our RALU is complementarily applicable to existing temporal acceleration methods and timestep-distilled models, leading to up to 15.9$\times$ speedup.
\end{abstract}

\blfootnote{* Authors contributed equally. $\dag$ Corresponding author.}

\vspace{-2em}
\section{Introduction}
\label{sec:Intro}
\begin{figure}[!t]
    \centering
    \begin{tikzpicture}[
        promptblock/.style={font=\footnotesize, align=center, text width=7cm},
        modelblock/.style={font=\footnotesize, align=center, text width=3cm}]
        \node[inner sep=0] (image) {\includegraphics[width=0.97\linewidth]{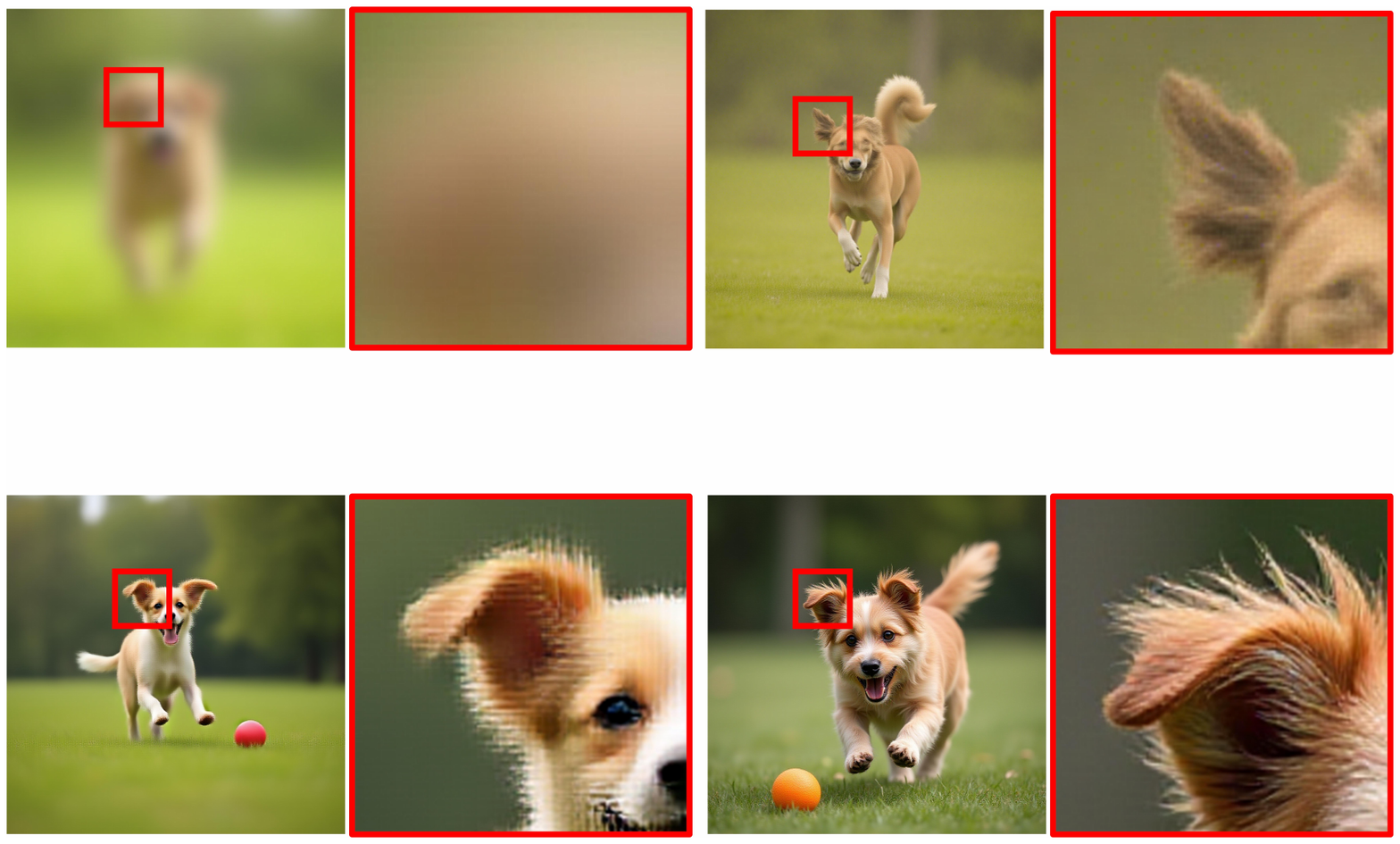}};
        \node[promptblock] at (0.1, -2.7)
        {\emph{``A dog is chasing after a ball and wagging its tail.''}};
        \node[modelblock] at (-2.0, 2.75)
        {FLUX-1.dev~\cite{black2024flux}\\(NFE = 7)};
        \node[modelblock] at (2.0, 2.75)
        {TaylorSeer~\cite{liu2025reusing}\\(Temporal accel.)};
        \node[modelblock] at (-2.0, -0.07)
        {Bottleneck Sampling~\cite{tian2025training}\\(Spatial accel.)};
        \node[modelblock] at (2.0, -0.07)
        {\textbf{RALU (Ours)}\\(Spatial accel.)};
    \end{tikzpicture}
    \caption{Generated 1024$\times$1024 images using acceleration methods on FLUX-1.dev~\cite{black2024flux} for 7$\times$ speedups.
    While temporal acceleration methods struggle with aggressive speedups and Bottleneck Sampling~\cite{tian2025training} introduces artifacts, our RALU successfully accelerates while avoiding artifacts and maintaining high image quality.
    }
    \label{fig:introduction}
    \vspace{-0.9em}
\end{figure}

Diffusion models~\cite{sohl2015deep,song2020denoising,ho2020denoising} have emerged as a dominant framework for generative modeling across diverse modalities such as image~\cite{rombach2022high,saharia2022photorealistic,nichol2021glide, chang2023muse, zhang2023adding}, video~\cite{ho2022video,blattmann2023align,liu2024sora, hongcogvideo, chen2024videocrafter2} and audio~\cite{guo2024audio,liu2023audioldm, schneider2024mousai}. Convolutional U-Net architectures~\cite{ronneberger2015u} have served as the backbone of relatively small diffusion models. Recently, diffusion transformers (DiTs)~\cite{rombach2022high,chen2024pixart,liu2024sora,black2024flux} have been introduced, leveraging the scalability of transformers to achieve state-of-the-art results in visual synthesis.

Despite the superior performance of DiTs, they suffer from high inference latency. This computational bottleneck is inherent in their transformer-based architecture, whose complexity scales quadratically with the number of input tokens~\cite{peebles2023scalable, lou2024token}. As models scale to higher resolutions and process more tokens, computational cost and latency pose a great challenge for their deployment in real-world applications such as real-time interactive editing or on-device generation~\cite{li2023snapfusion}.


Distillation~\cite{feng2024relational, zhang2024distill} and post-training~\cite{ma2024learning, yuan2024ditfastattn, you2024layer} have been explored to tackle this challenge, but they require additional training and substantial computational cost. Training-free temporal acceleration has emerged as an alternative by reducing computation across timesteps via feature caching~\cite{deltadit, zou2024accelerating, yuan2024ditfastattn, liu2025region}. In contrast, spatial acceleration, which involves transitioning between resolutions for a quadratic reduction in computation,remains underexplored, particularly for training-free latent upsampling in DiT.

Prior works on spatial upsampling, often explored within the tasks such as super-resolution~\cite{zheng2024self, zhou2024upscale} and higher-resolution synthesis~\cite{kim2024beyondscene, gao2025lumina, podell2023sdxl}, have focused on image space rather than latent space.
There are only a few existing works for upsampling on the latent space, but mostly require extra training, limiting their applicability to existing generative models that were heavily trained. 
Pyramidal Flow~\cite{jin2024pyramidal} 
requires the unified training for its generative framework. Similarly, Latent-SR~\cite{jeong2025latent} achieves latent space super-resolution by proposing a learned noise addition module. Bottleneck Sampling~\cite{tian2025training} is a training-free framework for spatial acceleration, but suffers from substantial artifacts, as illustrated in~\cref{fig:introduction} (bottom left).

Here, we investigate latent upsampling for DiTs. While late latent upsampling is beneficial for spatial acceleration, it introduces two types of artifacts: \textit{aliasing artifacts} in high-frequency regions and \textit{mismatching artifacts} from a noise-timestep discrepancy. We found that early upsampling mitigates aliasing artifacts, but requires more computation in high latent resolution. Then, we propose Region-Adaptive Latent Upsampling (RALU), a \textit{training-free}, \textit{spatial} acceleration approach for DiTs that efficiently and effectively exploits this trade-off. For RALU, we propose mixed-resolution latent upsampling that performs early upsampling only on artifact-prone regions and Noise and Timestep Matching (NT-Matching) that matches the noise distributions before and after upsampling. 
As illustrated in~\cref{fig:introduction}, RALU achieves computational gains by processing latents at low resolution mostly, while preserving high generation quality with significantly fewer artifacts by selectively performing early latent upsampling with NT-Matching. Moreover, RALU is complementarily applicable to existing temporal acceleration methods and timestep-distilled models, enabling additional speedups with negligible degradation in quality.
The contributions of this work are summarized as follows:
\begin{itemize}
    \item We investigate latent upsampling for DiTs as a means of spatial acceleration and explore the trade-off of upsampling between speedup and artifacts.
    \item We propose a training-free spatial acceleration method, RALU, that effectively exploits the trade-off by our mixed-resolution latent upsampling with early upsampling on artifact-prone edge regions and our NT-Matching for different latent resolutions, leading to up to 7.0$\times$ speedup with negligible quality degradation.
    \item We demonstrate that our RALU can be synergistically applicable to existing temporal acceleration methods and timestep-distilled models, leading to up to 15.9$\times$ speedup.
\end{itemize}

\section{Related Works}
\subsection{Flow matching}
Flow matching~\cite{lipman2022flow} is a recent generative model that learns a deterministic transport map from a simple prior (\textit{e.g.}, standard Gaussian noise) to a complex data distribution by integrating an ordinary differential equation (ODE).
In particular, rectified flow~\cite{liu2022flow} defines a linear interpolation path between the noise $\mathbf{x}_0$ and the data sample $\mathbf{x}_1$ by
\begin{equation}
\label{eq:rectified_flow}
\mathbf{x}_t = (1 - t)\mathbf{x}_0 + t\mathbf{x}_1, \quad t \in [0, 1],
\end{equation}
with a constant velocity field
$
\mathbf{v}_t = \frac{d\mathbf{x}_t}{dt} = \mathbf{x}_1 - \mathbf{x}_0.
$
The learning objective is to train a neural network
that takes the state $\mathbf{x}_t$ and time $t$ as input 
to predict this ground-truth conditional velocity field by minimizing the difference between the predicted velocity and the true velocity.

\subsection{Diffusion Transformer acceleration}
DiTs are computationally expensive, especially when generating high-resolution images, as the cost of self-attention grows quadratically with the number of spatial tokens. To mitigate these bottlenecks, recent research has proposed various inference-time acceleration techniques, which can be broadly categorized into model compression, temporal acceleration, and spatial acceleration methods.

\paragraph{Model compression.}
Model compression techniques aim to reduce model size or computational complexity without retraining from scratch.
Common approaches include quantization~\cite{shang2023post, li2024svdqunat, chen2024q}, distillation~\cite{li2023snapfusion, feng2024relational, zhang2024distill}, and block pruning~\cite{fang2024tinyfusion,xie2025sana,ma2024learning,seo2025skrr}. In particular, block pruning methods skip transformer blocks that contribute less during inference, improving efficiency. However, they often require a healing strategy, as quality degradation occurs without fine-tuning.


\paragraph{Temporal acceleration.}
Temporal acceleration aims to reduce computation by skipping layers or reusing cached features across timesteps.
Caching-based approaches have been extended to DiTs by storing internal activations such as block outputs~\cite{deltadit,ma2024deepcache,zou2024accelerating, liu2025timestep} or attention maps~\cite{yuan2024ditfastattn}.
Other strategies move beyond reuse, instead forecasting future features by modeling their temporal trajectory~\cite{liu2025reusing}.
Some works explore token-level pruning or selective execution~\cite{liu2025region}, or introduce learnable token routers that dynamically decide which tokens to recompute and which to reuse~\cite{you2024layer,lou2024token}.

\paragraph{Spatial acceleration.} 
Spatial acceleration refers to reducing computation by dynamically transitioning between latent resolutions.
In DiTs, this is equivalent to adjusting the number of input patches (tokens).
Recent studies~\cite{saharia2022photorealistic, ho2022cascaded, teng2023relay, jin2024pyramidal} have proposed cascaded diffusion frameworks that start from low-resolution and achieve high-resolution through upsampling during the denoising process.
However, these frameworks require resource-intensive training.

To the best of our knowledge, Bottleneck Sampling~\cite{tian2025training} is the only existing training-free spatially accelerated image generation method.
It achieves acceleration through two latent resolution changes via Lanczos resampling, but does not seem to correct for the skewed trajectory after upsampling, resulting in substantial artifacts. This highlights the need for a spatial acceleration method that can effectively mitigate such artifacts due to latent upsampling.

\section{Challenges in Spatial Acceleration for DiTs}
\label{sec:artifacts}

Multi-resolution approaches have been popular for many vision tasks, leading to spatial acceleration. However, while latent upsampling for DiTs has a great potential for spatial acceleration, cutting computational cost, it has not been well-investigated. In this section, we argue that na\"ive latent upsampling for acceleration introduces two types of artifacts in DiTs and describe our key findings on how each type of artifact can be effectively mitigated. 

\subsection{Aliasing artifacts due to late upsampling}
\label{subsec:artifacts}



\begin{figure}[t]
  \centering
  \begin{subfigure}{0.23\linewidth}
  \includegraphics[width=1.0\linewidth]{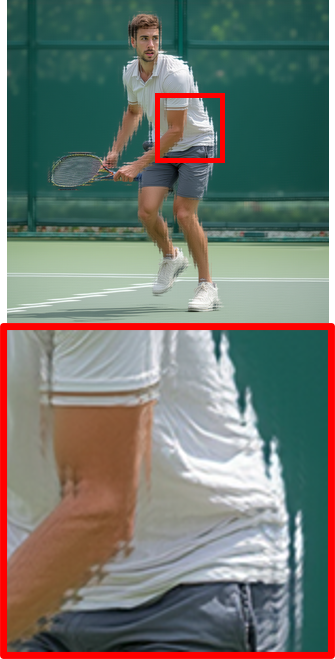}
    \caption
    {Aliasing.}
    \label{fig:aliasing_example}
    \end{subfigure}
  \hspace{0.3cm}
  \begin{subfigure}{0.65\linewidth}
  \includegraphics[width=1.0\linewidth]{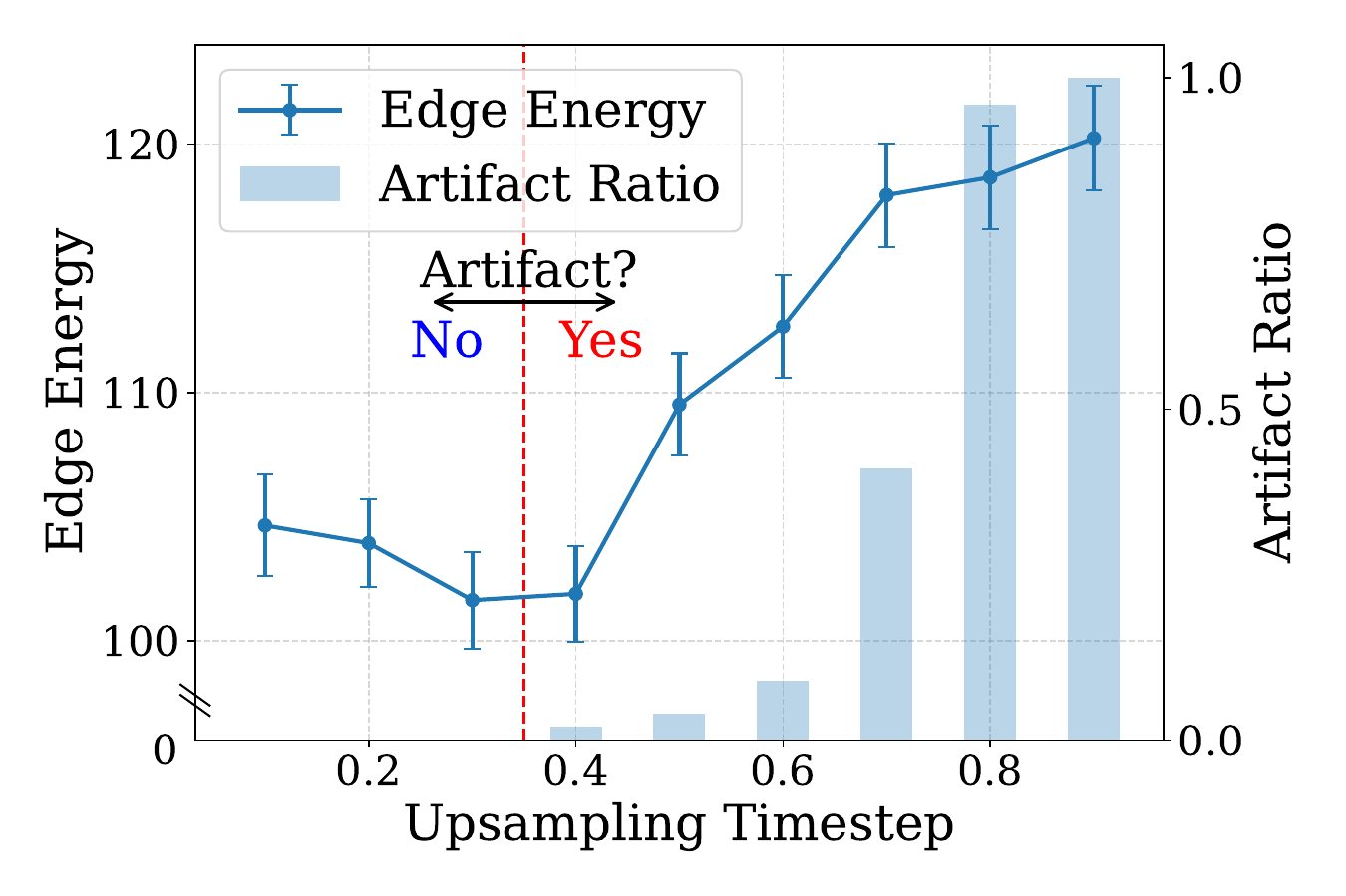}
    \caption
    {Analysis on upsampling timestep and artifact.}
    \label{fig:analysis_aliasing}
    \end{subfigure}
  \caption{(a) An example of aliasing artifacts generated using FLUX-1.dev (prompt: \emph{``A man on the tennis court is about to use his racket''}) with 9 low-resolution steps, 2$\times$ upsampling, and 9 full-resolution steps.
  (b) Edge energy and aliasing artifact ratio over image vs. upsampling timestep, averaged over 100 images.
  }
\end{figure}

We observe that various latent upsampling methods (\textit{e.g.}, nearest-neighbor, bilinear, bicubic, Lanczos) for DiT-based generations introduce aliasing artifacts predominantly in high-frequency regions such as semantic edges when upsampled at late timesteps.
Since the low-resolution latent space may not have enough representation power for high-frequency details such as sharp boundaries, late upsampling often leads to visual distortions. \cref{fig:aliasing_example} visualizes this phenomenon: while fidelity is maintained within smooth interior regions, aliasing artifacts appear prominently near boundaries when upsampled at the mid-timestep among 18 steps.

We found that these artifacts can be avoided by upsampling in earlier diffusion timesteps when the generated semantic structure is still coarse. To empirically validate this, we conducted the experiment with two-stage processes: denoising diffusion in low-resolution latent space and restoring full-resolution in various timesteps $t_{up}$. \cref{fig:analysis_aliasing} illustrates that upsampling in the early timestep ($t_{up} \leq 0.3$) does not introduce aliasing artifacts, while late upsampling ($t_{up} \geq 0.5$) leads to a significantly higher aliasing artifact ratio (i.e., the proportion of images exhibiting visible aliasing) and their undesirable edge energy (i.e., the average edge intensity quantified by a Canny edge detector).

\begin{remark}
    Aliasing artifacts in DiTs occur mostly in edge regions when upsampling latents at later timesteps, but do not occur when upsampling latents at earlier timesteps.
\end{remark}

However, na\"ively upsampling all latents in the early stages would sacrifice the computational benefits at a lower resolution. Due to this trade-off, mixed-resolution latent upsampling is desirable, which performs early upsampling only on the edge-region latent and late upsampling on the rest of the latents for mitigating aliasing artifacts and spatially accelerating image generation in DiTs, respectively. 

\begin{figure}[t]
  \centering
  \begin{subfigure}{0.23\linewidth}
  \includegraphics[width=0.99\linewidth]{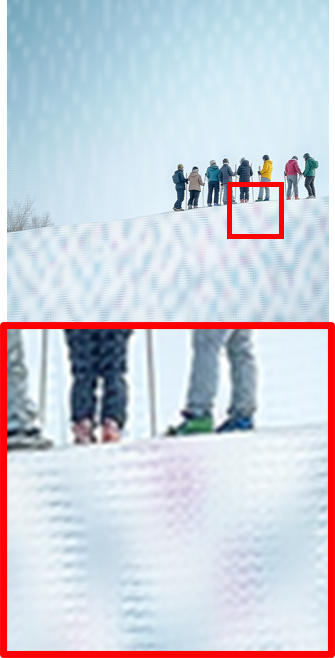}
    \caption
    {Mismatching.}
    \label{fig:mismatch_example}
    \end{subfigure}
  \hspace{0.3cm}
  \begin{subfigure}{0.65\linewidth}
  \includegraphics[width=0.99\linewidth]{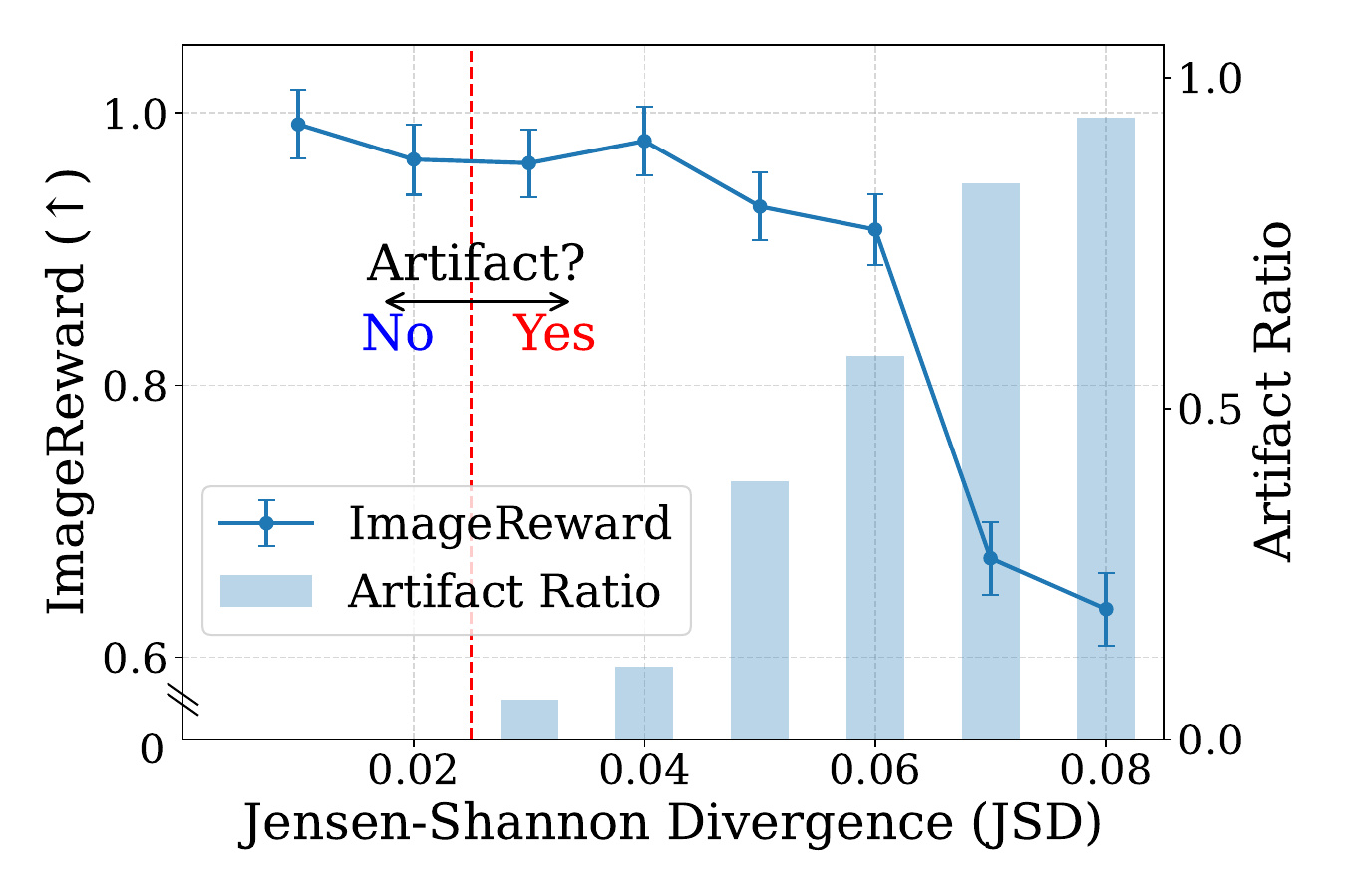}
    \caption
    {Analysis on JSD and performance.}
    \label{fig:analysis_mismatch}
    \end{subfigure}
  \caption{(a) An example of mismatching artifacts generated using FLUX-1.dev (prompt: \emph{''A group of people standing on top of a snow covered ski slope''}) with early upsampling ($t_{up} = 0.3$) and noise injection.
  (b) ImageReward~\cite{xu2023imagereward} score and mismatching artifact ratio vs. JSD, averaged over 100 images.
  }
\end{figure}

\subsection{Distribution mismatching artifacts}
\label{subsec:noise_level}

\begin{figure*}[!t]
  \centering
  \includegraphics[width=\linewidth]{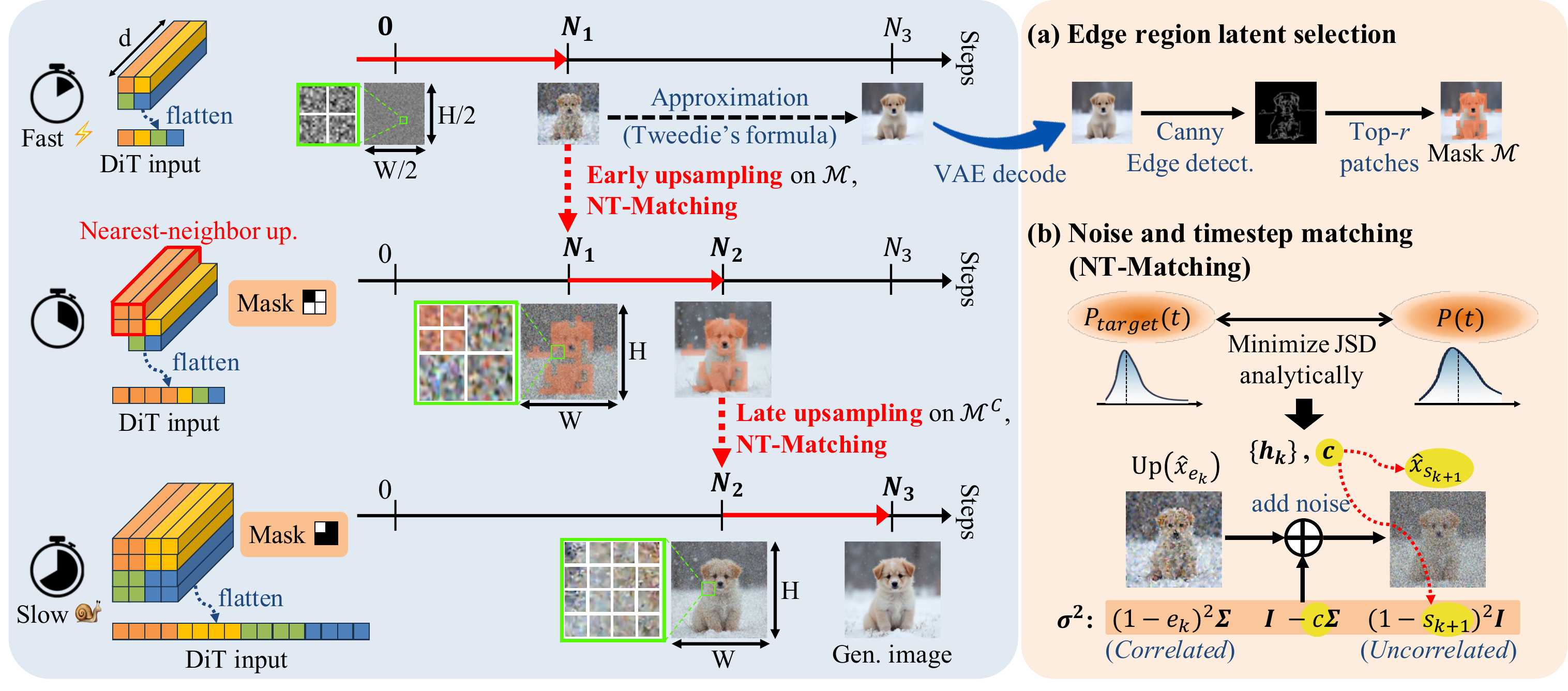}
    \caption{Overview of the proposed RALU framework. RALU consists of three different resolution processes: (1) low-resolution sampling for early denoising, (2) mixed-resolution sampling by upsampling edge region latents, and (3) full-resolution refinement by upsampling all remaining latents.
    (a) We select the top $r$ fraction of patches with the strongest edge signals from the decoded image and upsample them early.
    (b) We add correlated noise to the upsampled latents and design a corresponding timestep schedule.
    (See~\cref{subsec:NT-rescheduling} for more details).}
    \label{fig:overall_framework}
\end{figure*}

Latent upsampling for flow-based generations introduces distribution mismatching artifacts without properly injecting noise and matching timestep distribution (i.e., how frequently timesteps are sampled during inference~\cite{zheng2024beta}).

\paragraph{Correlated noise after upsampling.}
After upsampling, latent distribution deviates from the original flow trajectory. Starting from an initial noise $\mathbf{x}_0 \sim \mathcal{N}(0, \mathbf{I})$, flow matching aims to learn a mapping to the target $\mathbf{x}_1$ by following Eq.~\eqref{eq:rectified_flow} through a denoising process. Then, the conditional distribution of $\hat{\mathbf{x}}_t$ at timestep $t$ is:
\begin{equation}
\hat{\mathbf{x}}_t | \mathbf{x}_1\sim \mathcal{N}\left(t\mathbf{x}_1, (1-t)^2 \mathbf{I}\right).
\label{eq:conditional_distribution}
\end{equation}
After upsampling, the distribution becomes:
\begin{equation}
\text{Up}(\hat{\mathbf{x}}_{t}) | \mathbf{x}_1 \sim \mathcal{N}\left(t\text{Up}(\mathbf{x}_1), (1-{t})^2 \mathbf{\Sigma}\right),
\label{eq:upsampled_conditional_distribution}
\end{equation}
where $\text{Up}(\cdot)$ is a upsampling function and $\mathbf{\Sigma}$ is the covariance matrix after upsampling. For any interpolation-based upsampler, $\text{Up}(\hat{\mathbf{x}}_{t})$ no longer matches the original trajectory's distribution due to non-isotropic $\mathbf{\Sigma}$. To enforce $\text{Up}(\hat{\mathbf{x}}_{t})$ stay in the original trajectory, correlated noise should be injected, making the upsampled covariance matrix isotropic.


\paragraph{Timestep distribution matching.}
Noise injection alters noise level, which in turn skews timestep distribution. Models can be trained for this modification~\cite{teng2023relay, jin2024pyramidal}, but additional massive computation is required for each model. The mismatch between the pretrained model's original timestep distribution and the modified distribution seems to cause oversampling certain intervals and a frequency imbalance, resulting in visual artifacts.
\cref{fig:mismatch_example} visualizes this: early upsampling avoids aliasing, but the lack of distribution matching creates global mismatching artifacts.

We found that these artifacts can be mitigated by aligning the upsampling-induced timestep distribution with the original model's.
To verify this, we constructed a two-stage framework with fixed upsampling timestep to $t_{up}=0.3$ (to prevent aliasing) and applied our corrective noise and timestep distribution (detailed in~\cref{subsec:NT-rescheduling}) while varying its parameters to induce a range of Jensen-Shannon divergence (JSD) values. \cref{fig:analysis_mismatch} shows that as JSD decreases, mismatching artifacts diminish (disappearing at JSD $<0.03$) and the ImageReward~\cite{xu2023imagereward} score improves $(p<0.001)$.


\begin{remark}
Mismatching artifacts in flow-based DiTs occur when na\"ively adding noise after upsampling, but can be mitigated by aligning the new sampling distribution with the original timestep distribution of the pretrained model.
\end{remark}

\section{Region-Adaptive Latent Upsampling}

To break the fundamental trade-off between acceleration and artifacts identified in~\cref{sec:artifacts}, we propose Region-Adaptive Latent Upsampling (RALU) approach.

\subsection{Mixed-resolution latent upsampling}
\label{subsec:RALU}

As illustrated in~\cref{fig:overall_framework}, our approach comprises
mixed-resolution latent upsampling.
The denoising process begins at \textit{low-resolution} to accelerate. We reduce the latent resolution by a factor of 2 along width and height, thereby reducing the number of latent tokens to only 1/4.

Next, we perform an intermediate \textit{edge region upsampling} step to prevent aliasing artifacts, which typically occur in high-frequency regions as discussed in~\cref{subsec:artifacts}. We identify artifact-prone regions by 1) estimating the clean latent $\mathbf{x}_0$ via Tweedie’s formula from the low-resolution stage, 2) decoding it with the VAE, and 3) performing Canny edge detection. We then select and upsample the top-$r$ ratio of latent patches corresponding to these edges. The variable $r$ controls the trade-off between acceleration and artifact and we chose it in the range of 20-30\% of the whole latents, making the mixed-resolution latents.

\begin{figure}[t]
  \centering
  \includegraphics[width=0.99\linewidth]{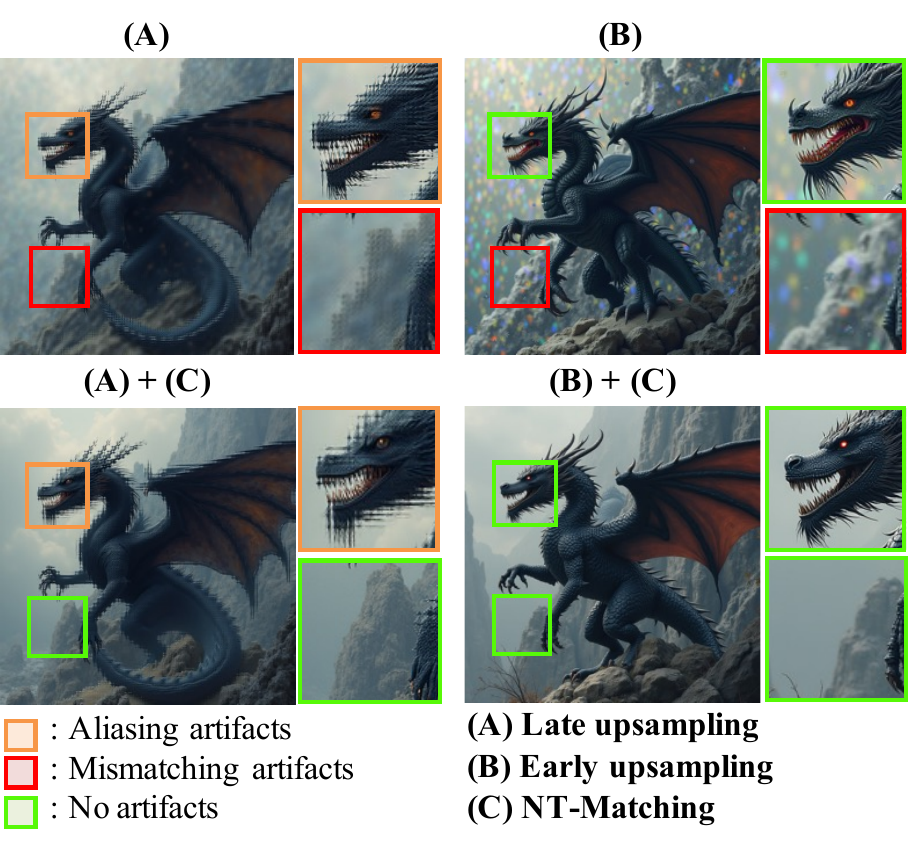}
    \caption
    {Resolving artifacts from na\"ive latent upsampling. \textcolor{orange}{Aliasing artifacts} are avoided by (B) early upsampling, while \textcolor{red}{Mismatching artifacts} are mitigated by (C) noise and timestep matching.}
    \label{fig:artifacts}
\end{figure}

Finally, the remaining low-resolution latent tokens are upsampled to {full-resolution} to generate a complete high-resolution image. This final step ensures the consistency between the early and late upsampled regions in the final output by matching position embeddings.
As demonstrated in~\cref{fig:artifacts}, late upsampling (left) results in aliasing, while early upsampling (right) effectively avoids this artifact.

\subsection{Noise and timestep matching}
\label{subsec:NT-rescheduling}
As discussed in~\cref{subsec:noise_level}, we design appropriate correlated noise injection and timestep distribution matching after latent upsampling to mitigate mismatching artifacts in DiTs.

\paragraph{Noise injection.}
Starting from Eq.~\eqref{eq:upsampled_conditional_distribution}, at the last timestep of the $k$-th stage $e_k$, the conditional distribution of the latent after 2$\times$ nearest-neighbor upsampling (selected for simplicity;
see supplementary material for details) is:
\begin{equation}
\label{eq:endpoint_upsample}
\text{Up}(\hat{\mathbf{x}}_{e_k}) | \mathbf{x}_1 \sim \mathcal{N}\left(e_k\text{Up}(\mathbf{x}_1), (1-{e_k})^2 \mathbf{\Sigma}\right),
\end{equation}
where $\mathbf{\Sigma}$ has a block-diagonal structure with the 4$\times$4 blocks filled with ones. As discussed in~\cref{subsec:noise_level}, since $\mathbf{\Sigma}$ is non-isotropic, 
proper correlated noise $\mathbf{z} \sim \mathcal{N}(0, \mathbf{\Sigma'})$ should be added to put it back onto the original trajectory's distribution:
\begin{gather}
a\text{Up}(\hat{\mathbf{x}}_{e_k})+ b\mathbf{z} = \hat{\mathbf{x}}_{s_{k+1}}, \; \mathbf{z}\sim \mathcal{N}(0, \mathbf{\Sigma'})\\
\hat{\mathbf{x}}_{s_{k+1}}| \mathbf{x}_1 \sim \mathcal{N}\left(s_{k+1}\text{Up}(\mathbf{x}_1), (1-s_{k+1})^2 \mathbf{I}\right).
\label{eq:noise_injection}
\end{gather}
where $s_{k+1}$ is the starting timestep of stage $k+1$ and $a$, $b$ are scalar values,
and $\mathbf{\Sigma'}$ is the covariance matrix of $\mathbf{z}$.
The parameters $s_{k+1}$, $a$, and $b$ are derived by matching the 
distribution in Eq.~\eqref{eq:noise_injection} under the constraint $\mathbf{\Sigma'} = \mathbf{I} - c\mathbf{\Sigma}$. By defining the composite term $\delta_k$ as $\delta_k \equiv (1-e_k)/\sqrt{c}$,
the rescheduling parameters can be expressed as:
\begin{equation}
\label{eq:timestep_reschedule_inline}
s_{k+1} = \frac{e_k}{\delta_k + e_k}, ~ a = \frac{1}{\delta_k + e_k}, ~ \text{and} ~ b = \frac{\delta_k}{\delta_k + e_k}.
\end{equation}
Detailed derivations are provided in the supplementary material.
This noise injection method modifies the approach from the {training-based} method of~\cite{jin2024pyramidal}. While~\cite{jin2024pyramidal} can learn the noise and timestep parameters during its unified training, our \textit{training-free} setting necessitates a different solution, as these parameters cannot be arbitrarily determined for a pre-trained model. Therefore, we introduce a timestep distribution matching that is compatible with pretrained models.

\begin{table*}[t!]
    \centering
    \caption{Quantitative comparisons of RALU with the baselines on (Top) FLUX.1-dev~\cite{black2024flux} (FLUX) and (Bottom) Stable Diffusion 3~\cite{esser2024scaling} (SD3).
    Performance is evaluated with CLIP-IQA~\cite{wang2023exploring} and NIQE~\cite{mittal2012making} for image quality, T2I-CompBench~\cite{huang2023t2i} and GenEval~\cite{ghosh2023geneval} for image-text alignment, and ImageReward~\cite{xu2023imagereward} for both.
    The number in parentheses next to FLUX indicates the total number of inference steps.
    $\uparrow / \downarrow$ denotes that a higher / lower metric is favorable.
    Speedup (Speed.) is calculated relative to the base model FLOPs (floating-point operations).
    \textcolor{blue}{\textbf{T}} and \textcolor{red}{\textbf{S}} denote the temporal and spatial acceleration, respectively.
    Additional computational metrics are reported in the supplementary material.
    }
    \label{tab:quantitative_flux}
    \resizebox{\textwidth}{!}{
    \begin{scriptsize}
    \begin{tabular}{cccccccccc}
         \toprule
        \multirow{2}{*}{Method} & \multirow{2}{*}{Accel.} & \multirow{2}{*}{Latency (s) $\downarrow$} & \multirow{2}{*}{TFLOPs $\downarrow$} & \multirow{2}{*}{Speed. $\uparrow$} & Overall & \multicolumn{2}{c}{Image quality} & \multicolumn{2}{c}{Text alignment} \\
        \cmidrule(lr){6-10}
        & & & & & ImageReward $\uparrow$ & CLIP-IQA $\uparrow$ & NIQE $\downarrow$ &T2I-Comp. $\uparrow$ & GenEval $\uparrow$  \\
        \midrule
        FLUX (50) & - & 25.1 & 2991.0 & 1.00$\times$ & 1.095 & 0.707 & 6.75 & 0.634 & 0.698  \\
        \cmidrule(lr){1-10}
        FLUX (10) & \textcolor{blue}{\textbf{T}} & 5.18 & 610.02 & 4.90$\times$ & \cellcolor{tabsecond}0.981 & \cellcolor{tabthird}0.679 & \cellcolor{tabthird}6.93 & 0.618 & \cellcolor{tabthird}0.647 \\
        $\Delta$-DiT~\cite{deltadit}  & \textcolor{blue}{\textbf{T}} & 7.42 & 772.10 & 3.87$\times$ & 0.102 & 0.487 & 9.60 & 0.306 & 0.397 \\
        ToCa~\cite{zou2024accelerating} & \textcolor{blue}{\textbf{T}} & 15.5 & 601.12 & 4.98$\times$ & -1.827 & 0.253 & 10.6 & 0.259 & 0.137 \\
        TeaCache~\cite{liu2025timestep}  & \textcolor{blue}{\textbf{T}} & 5.23 & 610.59 & 4.90$\times$ & 0.944 & 0.665 & 7.92 & \cellcolor{tabsecond}0.620 & \cellcolor{tabthird}0.647 \\
        TaylorSeer~\cite{liu2025reusing} & \textcolor{blue}{\textbf{T}} & 9.34 & 556.72 & 5.37$\times$ & \cellcolor{tabthird}0.972 & \cellcolor{tabsecond}0.684 & \cellcolor{tabsecond}6.77 & 0.594 & 0.619 \\
        Bottleneck~\cite{tian2025training} & \textcolor{red}{\textbf{S}} & 5.37 & 571.23 & 5.24$\times$ & 0.889 & 0.661 & 9.16 & \cellcolor{tabsecond}0.620 & \cellcolor{tabfirst}0.687 \\
        \textbf{RALU (Ours)} & \textcolor{red}{\textbf{S}} & \textbf{5.04} & \textbf{540.47} & \textbf{5.53$\times$} & \cellcolor{tabfirst}1.022 & \cellcolor{tabfirst}0.700 & \cellcolor{tabfirst}6.43 & \cellcolor{tabfirst}0.626 & \cellcolor{tabsecond}0.652 \\
        \cmidrule(lr){1-10}
        FLUX (7) & \textcolor{blue}{\textbf{T}} & 3.79 & 431.45 & 6.93$\times$ & \cellcolor{tabsecond}0.920 & \cellcolor{tabsecond}0.660 & \cellcolor{tabsecond}8.25 & 0.594 & 0.583 \\
        TeaCache~\cite{liu2025timestep} & \textcolor{blue}{\textbf{T}} & 4.21 & 431.83 & 6.93$\times$ & 0.733 & 0.623 & 13.7 & \cellcolor{tabthird}0.599 & \cellcolor{tabthird}0.594 \\
        TaylorSeer~\cite{liu2025reusing} & \textcolor{blue}{\textbf{T}} & 7.00 & 431.74 & 6.83$\times$ & 0.660 & \cellcolor{tabthird}0.646 & 9.43 & 0.514 & 0.446 \\
        Bottleneck~\cite{tian2025training} & \textcolor{red}{\textbf{S}} & 3.78 & 431.52 & 6.93$\times$ & \cellcolor{tabthird}0.792 & 0.631 & \cellcolor{tabthird}8.71 & \cellcolor{tabsecond}0.605 & \cellcolor{tabsecond}0.672 \\
        \textbf{RALU (Ours)} & \textcolor{red}{\textbf{S}} & \textbf{3.75} & \textbf{426.01} & \textbf{7.02$\times$} & \cellcolor{tabfirst}0.999 & \cellcolor{tabfirst}0.681 & \cellcolor{tabfirst}6.87 & \cellcolor{tabfirst}0.633 & \cellcolor{tabfirst}0.682 \\
        \bottomrule
    \end{tabular}
    \end{scriptsize}
    }
    \resizebox{\textwidth}{!}{
    \begin{scriptsize}
    \begin{tabular}{cccccccccc}
         \toprule
        \multirow{2}{*}{Method} & \multirow{2}{*}{Accel.} & \multirow{2}{*}{Latency (s) $\downarrow$} & \multirow{2}{*}{TFLOPs $\downarrow$} & \multirow{2}{*}{Speed. $\uparrow$} & Overall & \multicolumn{2}{c}{Image quality} & \multicolumn{2}{c}{Text alignment} \\
        \cmidrule(lr){6-10}
        & & & & & ImageReward $\uparrow$ & CLIP-IQA $\uparrow$ & NIQE $\downarrow$ &T2I-Comp. $\uparrow$ & GenEval $\uparrow$  \\
        \midrule
        SD3 (28) & - & 4.04 & 351.66 & 1.00$\times$ & 0.971 & 0.692 & 6.09 & 0.667 & 0.703  \\
        \cmidrule(lr){1-10}
        SD3 (14) & \textcolor{blue}{\textbf{T}} & 2.14 & 183.30 & 1.92$\times$ & 0.888 & \cellcolor{tabsecond}0.667 & 6.06 & 0.651 & 0.628  \\
        $\Delta$-DiT~\cite{deltadit} & \textcolor{blue}{\textbf{T}} & 2.41 & 183.46 & 1.92$\times$ & 0.875 & \cellcolor{tabsecond}0.667 & 5.87 & \cellcolor{tabsecond}0.660 & 0.640\\
        ToCa~\cite{zou2024accelerating} & \textcolor{blue}{\textbf{T}} & 2.62 & 181.51 & 1.94$\times$ & 0.885 & 0.663 & 5.72 & \cellcolor{tabthird}0.655 & \cellcolor{tabsecond}0.653\\
        RAS~\cite{liu2025region} & \textcolor{blue}{\textbf{T}} & 2.07 & 185.14 & 1.90$\times$ & 0.706 & 0.574 & \cellcolor{tabfirst}4.85 & 0.649 & 0.583 \\
        TaylorSeer~\cite{liu2025reusing} & \textcolor{blue}{\textbf{T}} & 2.40 & 183.48 & 1.92$\times$ & \cellcolor{tabsecond}0.913 & 0.664 & \cellcolor{tabthird}5.53 & 0.654 & \cellcolor{tabfirst}0.666 \\
        Bottleneck~\cite{tian2025training} & \textcolor{red}{\textbf{S}} & 2.14 & 186.21 & 1.89$\times$ & \cellcolor{tabthird}0.890 & 0.636 & 5.70 & 0.645 & 0.644 \\
        \textbf{RALU (Ours)} & \textcolor{red}{\textbf{S}} & \textbf{2.04} & \textbf{181.09} & \textbf{1.94$\times$} & \cellcolor{tabfirst}0.971 & \cellcolor{tabfirst}0.684 & \cellcolor{tabsecond}5.17 & \cellcolor{tabfirst}0.671 & \cellcolor{tabthird}0.647 \\
        \cmidrule(lr){1-10}
        SD3 (9) & \textcolor{blue}{\textbf{T}} & 1.46 & 123.17 & 2.86$\times$ & \cellcolor{tabthird}0.565 & \cellcolor{tabsecond}0.619 & 6.10 & \cellcolor{tabthird}0.627 & \cellcolor{tabthird}0.561 \\
        RAS~\cite{liu2025region} & \textcolor{blue}{\textbf{T}} & 1.40 & 118.99 & 2.96$\times$ & 0.423 & 0.529 & \cellcolor{tabthird}5.70 & 0.613 & 0.470 \\
        TaylorSeer~\cite{liu2025reusing} & \textcolor{blue}{\textbf{T}} & 1.87 & 123.49 & 2.85$\times$ & 0.059 & 0.521 & 7.44 & 0.561 & \cellcolor{tabsecond}0.572 \\
        Bottleneck~\cite{tian2025training} & \textcolor{red}{\textbf{S}} & 1.44 & 127.35 & 2.76$\times$ & \cellcolor{tabsecond}0.628 & \cellcolor{tabthird}0.566 & \cellcolor{tabsecond}5.58 & \cellcolor{tabsecond}0.629 & 0.528 \\
        \textbf{RALU (Ours)} & \textcolor{red}{\textbf{S}} & \textbf{1.38} & \textbf{116.61} & \textbf{3.02$\times$} & \cellcolor{tabfirst}0.798 & \cellcolor{tabfirst}0.645 & \cellcolor{tabfirst}5.44 & \cellcolor{tabfirst}0.661 & \cellcolor{tabfirst}0.599 \\
        \bottomrule
    \end{tabular}
    \end{scriptsize}
    }
\end{table*}


\paragraph{Timestep distribution matching.}
After noise injection at the timestep $e_k$, the diffusion process resumes at $s_{k+1}$,
meaning reusing the original schedule can oversample the interval $[s_{k+1}, e_k]$.
As mentioned in~\cref{subsec:noise_level}, this misalignment causes artifacts.
Noise and Timestep Matching (NT-Matching) resolves these artifacts by matching the timestep distribution with the original scheduler, 
such as the non-uniform sampling employed by flow matching~\cite{esser2024scaling, black2024flux}.
Their probability density function (PDF) and truncated PDF are:
\begin{align}
f_h(t) = \frac{h}{(1+(h-1)t)^2}\;\;,0 \leq t \leq 1,&\\f_{h,s,e}(t) = \frac{f_h(t)}{F_h(e)-F_h(s)}\;\;,s \leq t \leq e,
\end{align}
where $h$ is a shifting parameter and $F_h(t)$ is a cumulative distribution function (CDF) of $f_h(t)$. Our method injects noise at the end of each stage, which requires additional denoising over the overlapping interval $[s_{k+1}, e_k]$. Therefore, timestep sampling within intervals $[0, 1], [s_2, e_1], \dots, [s_K, e_{K-1}]$ should follow $f_h(t)$. The overall sampling distribution can be written as a weighted sum of truncated PDFs:
\begin{align*}
\label{eq:target_distribution}
P_\text{target}(t) =
\frac{f_{h_{\text{ori}},0,1}(t) + \sum_{k=1}^{K-1}(e_k-s_{k+1})
    f_{h_{\text{ori}},s_{k+1},e_k}(t)}{1+\sum_{k=1}^{K-1}(e_k-s_{k+1})}
\end{align*}
where $h_{ori}$ is the shifting parameter of the base model and $K=3$ is the number of stages.
However, the actual sampling intervals are $[0, e_1], [s_2, e_2], \dots, [s_K, 1]$. Therefore, the target distribution should also be adjusted accordingly. We use a stage-wise shifting parameter $h_k$ to control the PDF in each interval. Assuming we sample $N_k - N_{k-1}$ timesteps in the $k$-th interval according to $f_{h_k,s_k,e_k}(t)$, the modified timestep distribution $P(t)$ is:
\begin{equation}
\label{eq:modified_distribution}
P(t) = \frac{\sum_{k=1}^{K} (N_k - N_{k-1}) f_{h_k,s_k,e_k}(t)}{N_K}.
\end{equation}


\begin{figure*}[t]
    \centering
    \begin{tikzpicture}[
        promptblock/.style={font=\footnotesize, align=center, text width=7cm},
        modelblock/.style={font=\small, align=center, text width=3cm},
        baseblock/.style={font=\normalsize, align=center, text width=3.5cm}]
        \node[inner sep=0] (image) {\includegraphics[width=0.99\linewidth]{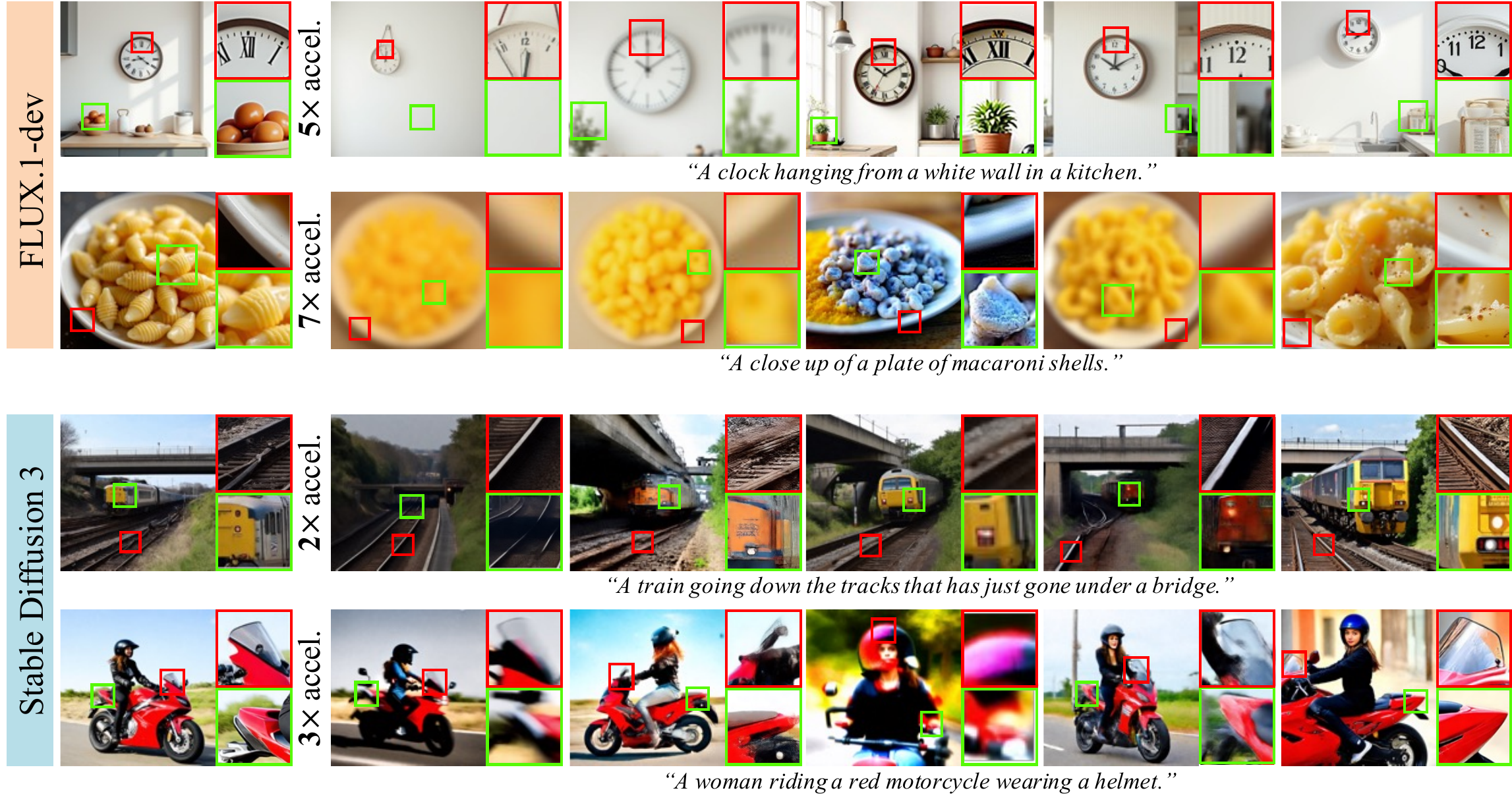}};
        \node[modelblock] at (-6.72, 4.8)
        {FLUX (NFE=50)};
        \node[modelblock] at (-3.6, 4.8)
        {FLUX (NFE=10,7)};
        \node[modelblock] at (-0.85, 4.8)
        {TeaCache~\cite{liu2025timestep}};
        \node[modelblock] at (1.89, 4.8)
        {TaylorSeer~\cite{liu2025reusing}};
        \node[modelblock] at (4.62, 4.8)
        {Bottleneck~\cite{tian2025training}};
        \node[modelblock] at (7.3, 4.8)
        {\textbf{RALU (Ours)}};

        \node[modelblock] at (-6.72, 0.05)
        {SD3 (NFE=28)};
        \node[modelblock] at (-3.6, 0.05)
        {SD3 (NFE=14,9)};
        \node[modelblock] at (-0.85, 0.05)
        {RAS~\cite{liu2025region}};
        \node[modelblock] at (1.89, 0.05)
        {TaylorSeer~\cite{liu2025reusing}};
        \node[modelblock] at (4.62, 0.05)
        {Bottleneck~\cite{tian2025training}};
        \node[modelblock] at (7.3, 0.05)
        {\textbf{RALU (Ours)}};

    \end{tikzpicture}
    \caption{Qualitative comparison of images generated by baseline methods and RALU on FLUX and SD3 under various speedups. For FLUX, we compare at 5$\times$ and 7$\times$ speedups; for SD3, at 2$\times$ and 3$\times$.
    NFE (number of function evaluations) refers to the number of inference steps.
    Zoomed-in regions on the right highlight that RALU preserves fine-grained details and avoids artifacts more effectively than the other baselines, even under high speedups. More results are provided in the supplementary material. Best viewed in zoom.
    }
    \label{fig:qualitative_results}
\end{figure*}

We minimize the Jensen-Shannon divergence between the target distribution $P_\text{target}(t)$ and the actual distribution $P(t)$ via numerical search.
We analytically determine $c$ and $\{h_{k}\}$, and these values then dictate the noise degree, correlation, and full timestep schedule. \cref{fig:artifacts} shows the effectiveness of NT-Matching: while the image with only early upsampling (top right) suffers from mismatching artifacts, adding NT-Matching (bottom right) successfully mitigates them.

\begin{table*}[t!]
\centering
\caption{
(Top) Quantitative results of integrating temporal acceleration methods into RALU under a 5$\times$ speedup setting on FLUX.
(Bottom) Quantitative results of adapting RALU on timestep-distilled models (FLUX.1-schnell, SD3.5L-Turbo). Speedups are measured relative to FLUX.1-dev and Stable Diffusion 3.5 Large, respectively.
\textcolor{green}{\textbf{D}} denotes the timestep-distilled model.
The $W$ value of TaylorSeer denotes the number of warm-up steps.
}
\resizebox{\textwidth}{!}{
    \begin{scriptsize}
    \begin{tabular}{ccccccccc}
         \toprule
        \multirow{2}{*}{Method} & \multirow{2}{*}{Accel.} & \multirow{2}{*}{TFLOPs $\downarrow$} & \multirow{2}{*}{Speed. $\uparrow$} & Overall & \multicolumn{2}{c}{Image quality} & \multicolumn{2}{c}{Text alignment} \\
        \cmidrule(lr){5-9}
        & & & & ImageReward $\uparrow$ & CLIP-IQA $\uparrow$ & NIQE $\downarrow$ & T2I-Comp. $\uparrow$ & GenEval $\uparrow$  \\
        \midrule
        \rowcolor{gray!20} RALU (5$\times$) & \textcolor{red}{\textbf{S}} & 540.47 & 5.53$\times$ & 1.022 & 0.700 & 6.43 & 0.626 & 0.652 \\
        \multicolumn{1}{l}{+ $\Delta$-DiT} & \textcolor{red}{\textbf{S}} + \textcolor{blue}{\textbf{T}} & 422.56 & 7.08$\times$ & 0.827 & 0.619 & 7.65 & 0.560 & 0.555\\
        \multicolumn{1}{l}{+ ToCa} & \textcolor{red}{\textbf{S}} + \textcolor{blue}{\textbf{T}} & 409.74 & 7.30$\times$ & 0.914 & 0.522 & 6.94 & 0.620 & 0.652\\
        \multicolumn{1}{l}{+ TaylorSeer ($W=3$)} & \textcolor{red}{\textbf{S}} + \textcolor{blue}{\textbf{T}} & 410.70 & 7.28$\times$ & 0.959 & 0.708 & 6.43 & 0.631 & 0.680 \\
        \multicolumn{1}{l}{+ TaylorSeer ($W=2$)} & \textcolor{red}{\textbf{S}} + \textcolor{blue}{\textbf{T}} & 331.38 & 9.03$\times$ & 0.926 & 0.691 & 6.00 & 0.590 & 0.586 \\
        \cmidrule(lr){1-9}
        \rowcolor{gray!20} FLUX.1-schnell (4) & \textcolor{green}{\textbf{D}} & 252.88 & 11.83$\times$ & 1.055 & 0.686 & 7.46 & 0.640 & 0.688 \\
        \multicolumn{1}{l}{+ RALU} & \textcolor{green}{\textbf{D}} + \textcolor{red}{\textbf{S}} & 187.92 & 15.91$\times$ & 0.992 & 0.650 & 6.88 & 0.621 & 0.636 \\
        \rowcolor{gray!20} SD3.5L-Turbo (4) & \textcolor{green}{\textbf{D}} & 107.93 & 6.31$\times$ & 0.840 & 0.670 & 7.20 & 0.596 & 0.744 \\
        \multicolumn{1}{l}{+ RALU} & \textcolor{green}{\textbf{D}} + \textcolor{red}{\textbf{S}} & 82.23 & 8.28$\times$ & 0.789 & 0.628 & 6.77 & 0.607 & 0.696 \\
        \bottomrule
    \end{tabular}
    \end{scriptsize}
}
\label{tab:combine_caching}
\end{table*}

\section{Experiments}
\subsection{Quantitative results}
\paragraph{Base models and metrics.}
We adopt FLUX.1-dev~\cite{black2024flux} (FLUX) and Stable Diffusion 3 Medium~\cite{esser2024scaling} (SD3) as base models. We assessed image quality and text alignment with ImageReward~\cite{xu2023imagereward}, CLIP-IQA~\cite{wang2023exploring}, NIQE~\cite{mittal2012making}, T2I-CompBench~\cite{huang2023t2i} and GenEval~\cite{ghosh2023geneval}. We measured acceleration using latency and FLOPs on a single A100 GPU.
More details are provided in the supplementary materials.

\paragraph{T2I generation performance comparison.}
We compare RALU with existing temporal acceleration methods such as $\Delta$-DiT~\cite{deltadit}, ToCa~\cite{zou2024accelerating}, TeaCache~\cite{liu2025timestep}, RAS~\cite{liu2025region}, TaylorSeer~\cite{liu2025reusing}, and the spatial acceleration method Bottleneck Sampling~\cite{tian2025training}.
\cref{tab:quantitative_flux} presents the results on FLUX (top) and SD3 (bottom).
Temporal acceleration methods struggle to deliver strong performance in both image quality and text alignment. The spatial acceleration baseline, Bottleneck Sampling, also suffers from degradation in image quality.
In contrast, RALU achieves higher image fidelity and text alignment compared to other baselines.

\subsection{Qualitative results}
As shown in~\cref{fig:qualitative_results}, base models with reduced inference steps and temporal acceleration methods tend to produce images with noticeable blur or artifacts.
Bottleneck Sampling also introduces aliasing or mismatching artifacts. In contrast, RALU outperforms both temporal and spatial acceleration baselines, maintaining superior visual fidelity and semantic alignment with minimal artifacts.

\subsection{Integrating RALU with other accelerations}
\label{subsec:caching}

As mentioned in~\cref{sec:Intro}, RALU is complementary to temporal acceleration and can be combined. \cref{tab:combine_caching} (Top) presents the quantitative results of incorporating temporal acceleration methods (\textit{e.g.}, caching, forecasting) into the RALU framework. This integration yields additional improvements in speed up to 9.03$\times$ without any training.

Furthermore, RALU is also compatible with timestep-distilled models. Unlike temporal acceleration, which is applied at inference without retraining, timestep distillation represents a training-based acceleration strategy that fine-tunes the model to operate efficiently in fewer steps. Despite this difference, RALU can be seamlessly integrated with such distilled models to provide further inference-time speedups. As demonstrated in~\cref{tab:combine_caching} (Bottom), this combination achieves up to a 15.91$\times$ total speedup over the base model with only minimal degradation in generation quality.
We also provide qualitative comparisons for these integrations in the supplementary material.

\begin{remark}
RALU can be integrated with temporal acceleration method or timestep-distilled model to achieve further training-free speedup while incurring only minimal quality degradation.
\end{remark}

\begin{figure*}[t]
\centering
\begin{minipage}{0.41\textwidth}
    \centering
    \includegraphics[width=0.99\linewidth]{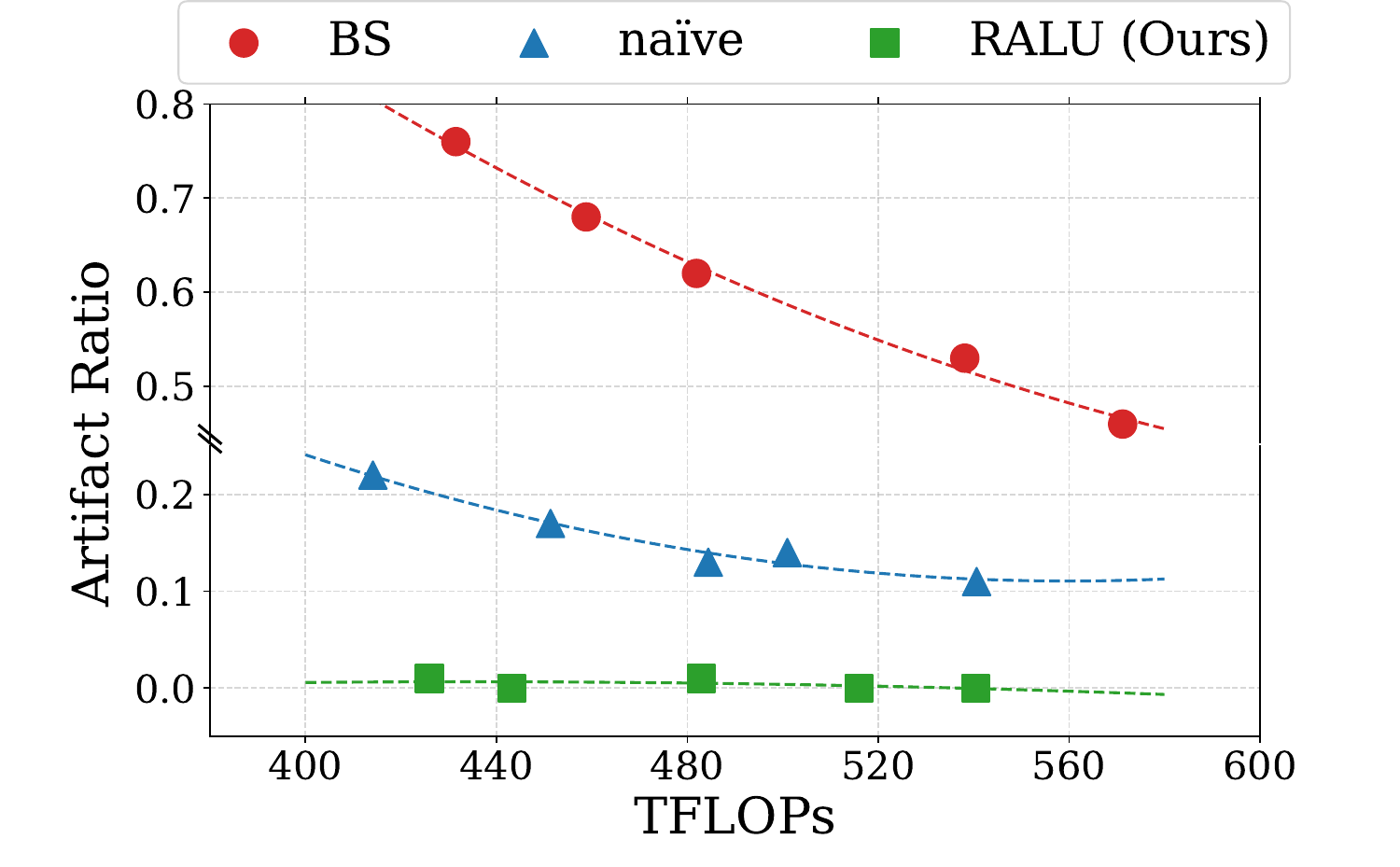}
    \captionsetup{type=figure} 
    \caption{RALU has a drastically lower artifact ratio than Bottleneck Sampling (BS) and a na\"ive upsampling for a similar TFLOPs. Trendlines represent quadratic fits.}
    \label{fig:ex_artifact}
\end{minipage}
\hspace{0.3cm}
\begin{minipage}{0.56\textwidth}
    \vspace{0.33cm}
    \centering
    \begin{subfigure}{0.49\linewidth}
        \includegraphics[width=1.0\linewidth]{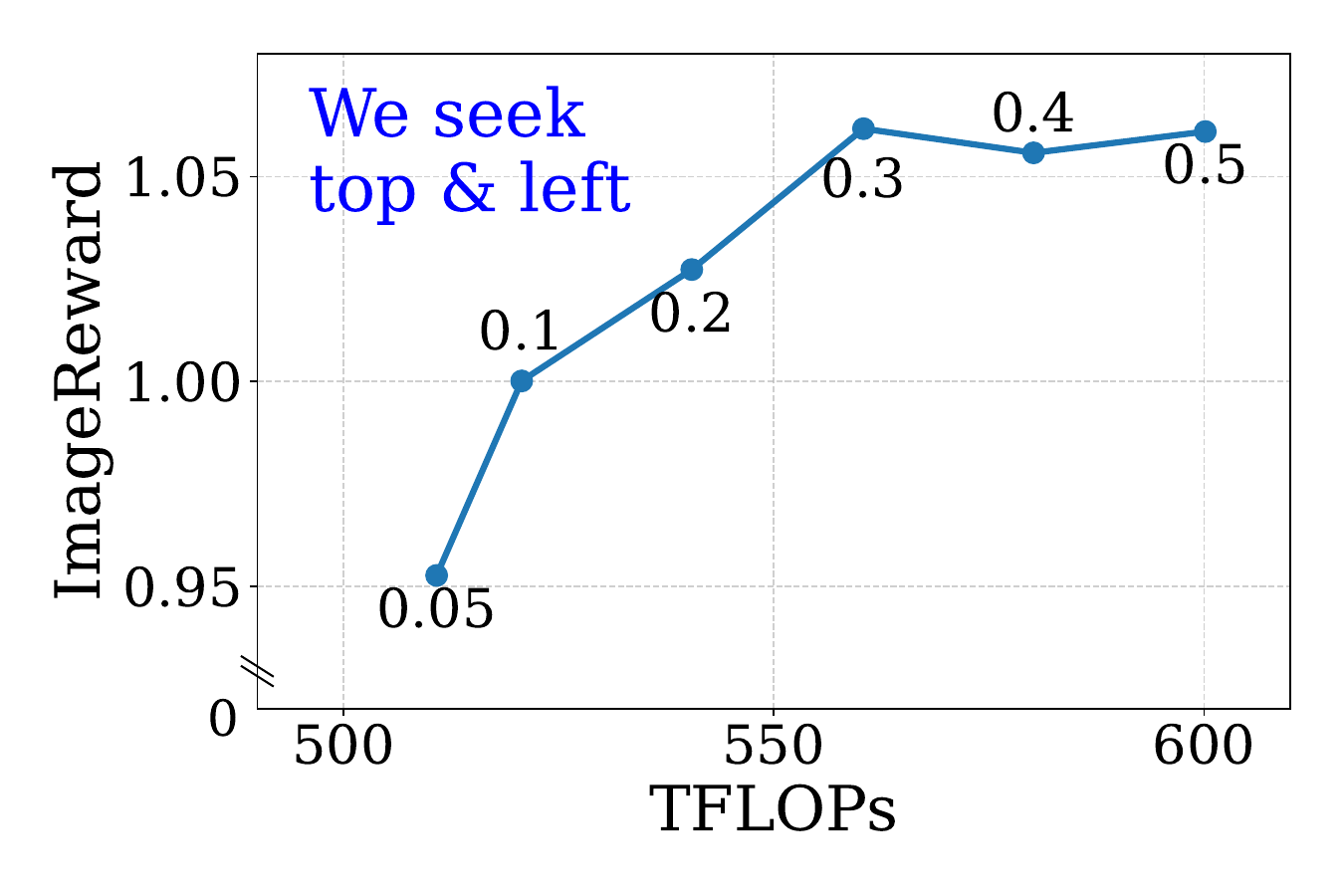}
        \caption{ImageReward vs. TFLOPs.}
        \label{fig:ablation_ratio_1}
    \end{subfigure}
    \begin{subfigure}{0.49\linewidth}
        \includegraphics[width=1.0\linewidth]{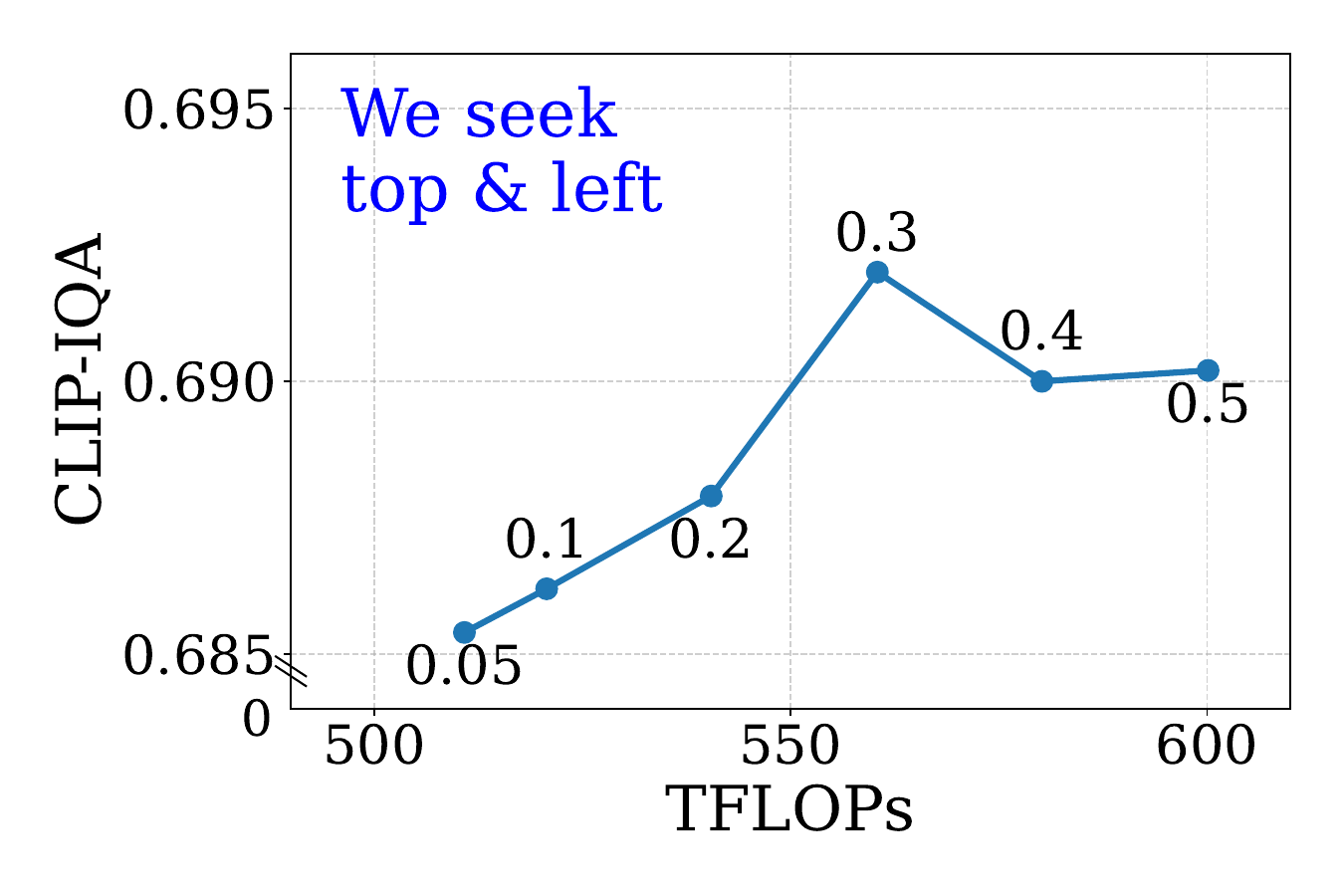}
        \caption{CLIP-IQA vs. TFLOPs.}
        \label{fig:ablation_ratio_2}
    \end{subfigure}
    \vspace{0.85cm}
    \captionsetup{type=figure}
    \caption{Trade-off between (a) ImageReward and (b) CLIP-IQA against TFLOPs for different upsampling ratios (0.05–0.5). Quality generally improves with the upsampling ratio but peaks around 0.3.}
    \label{fig:ablation_ratio}
\end{minipage}
\end{figure*}

\subsection{Artifact comparison}
\cref{fig:ex_artifact} shows the artifact ratio according to computational cost (TFLOPs).
Bottleneck Sampling~\cite{tian2025training} suffers from a high artifact ratio, and a na\"ive upsampling approach also results in a considerable artifact ratio. In contrast, our proposed RALU achieves a substantially lower artifact ratio than both baselines at similar TFLOPs for various speedup settings, demonstrating its superior artifact mitigation.

\begin{table}[t]
    \centering
    \caption{Effect of NT-Matching. Results are on the FLUX 7$\times$ setting. The optimized parameters yield JSD$=0.026$, while other parameters were adjusted to obtain higher JSDs for comparison.}
    \label{tab:ntdm}
    \resizebox{0.99\linewidth}{!}{
    \begin{footnotesize}
    \begin{tabular}{c|ccc}
        \toprule
        JSD & ImageReward $\uparrow$ & NIQE $\downarrow$ & T2I-CompBench $\uparrow$ \\
        \midrule
        0.026 (optimized) & \textbf{0.999} & \textbf{6.51} & \textbf{0.633} \\
        0.030 & 0.972 & 6.60 & 0.571 \\
        0.035 & 0.981 & 6.53 & 0.569 \\
        0.040 & 0.966 & 6.80 & 0.565 \\
        \bottomrule
    \end{tabular}
    \end{footnotesize}
    }
\end{table}

\subsection{Edge region selection}
\paragraph{Image vs. latent space.} For edge region selection, we apply a Canny edge detector in the image space. An alternative is to apply the Sobel operator directly in the latent space. As detailed in the supplementary material, we found that image-space detection yields more accurate edge maps and slightly better generative performance. We adopted the Canny detector as the additional VAE decoder overhead is modest (2.48 TFLOPs for FLUX and SD3).

\paragraph{Fixed vs. adaptive upsampling ratio.} We utilize a fixed top-$r$ ratio to select edge regions but also experimented with a strategy that adjusts this ratio dynamically. Results in the supplementary material show that, for a similar average FLOPs, the adaptive method did not noticeably improve average image fidelity over the fixed ratio. Given that the fixed ratio ensures consistent computational costs without sacrificing quality, we adopted it for its stability.

\subsection{Ablation study}

\paragraph{Effect of early upsampling ratio.}
\label{ablation:upsample_ratio}
In Stage 2 of RALU, increasing the amount of top-$r$ ratio of latent upsampling results in a more robust avoidance of aliasing artifacts. We define the upsampling ratio as the fraction of top-\(r\) ratio of latents selected for early upsampling. \cref{fig:ablation_ratio} highlights the impact of the upsampling ratio on the generated image quality. We note a trade-off where higher upsampling ratios improve image quality but lead to an increase in FLOPs.
Therefore, we adopt ratios in the 0.2-0.3 range, which offers the best compromise between quality and efficiency.

\paragraph{Effect of NT-Matching.}
We analyzed the effectiveness of NT-Matching by measuring metrics under various noise level and denoising timestep scenarios. As shown in~\cref{tab:ntdm}, it is evident that optimizing JSD yields the best image quality and text alignment, highlighting the superiority of NT-Matching.

\section{Discussion}
The precise mechanism by which diffusion models handle simultaneous sampling or denoising across varying resolutions, particularly how self-attention weights are computed in a multi-resolution setup, remains an area for further investigation. However, the observed robustness of these models to changes in input structure, such as the successful application of methods like token merging~\cite{bolya2023token, bolya2023token2, seo2025geometrical}, where multiple tokens are compressed into a single token without major performance drops, suggests an inherent capacity for resolution-agnostic or adaptive representation learning.

\section{Conclusion}
In this work, we proposed \textit{Region-Adaptive Latent Upsampling} (RALU), a training-free framework to accelerate Diffusion Transformers.
RALU resolves the trade-off between acceleration and aliasing artifacts in latent upsampling through its three-stage process: low-resolution denoising for acceleration, edge-selective upsampling to avoid aliasing, and full-resolution refinement.
Furthermore, to prevent mismatching artifacts, we introduced a noise-timestep matching scheme.
Experiments show that RALU achieves speedup with negligible quality loss and artifacts, and complements temporal acceleration methods for further gains, offering an effective solution for efficient DiT inference.

\section{Acknowledgements}
This work was supported in part by Institute of Information \& communications Technology Planning \& Evaluation (IITP) grant funded by the Korea government(MSIT) [NO.RS-2021-II211343, Artificial Intelligence Graduate School Program (Seoul National University)], the National Research Foundation of Korea(NRF) grants funded by the Korea government(MSIT) (Nos. RS-2025-02263628, RS-2022-NR067592) and Samsung Electronics MX Division. Also, the authors acknowledged the financial support from the BK21 FOUR program of the Education and Research Program for Future ICT Pioneers, Seoul National University.



\appendix
\renewcommand{\thefigure}{S\arabic{figure}}
\renewcommand{\thetable}{S\arabic{table}}
\renewcommand{\theequation}{S\arabic{equation}}
\setcounter{figure}{0}
\setcounter{table}{0}
\setcounter{equation}{0}


\section*{Appendix}

\section{Details on The Derivations}
\subsection{Derivation of Eq.~(8)}
\label{append:derivation_timestep}
Starting from Eq.~(4), the conditional distribution of the linear combination of upsampled latent $\text{Up}(\hat{\mathbf{x}}_{e_k})$ and the correlated noise $\mathbf{z} \sim \mathcal{N}(0, \mathbf{\Sigma'})$ is:
\begin{equation}
\label{append_eq:s1}
\begin{split}
\left( a\text{Up}(\hat{\mathbf{x}}_{e_k}) + b\mathbf{z} \right) | \mathbf{x}_1 \\ \sim \mathcal{N}\Biggl( & ae_k\text{Up}(\mathbf{x}_1),   a^2(1-{e_k})^2\mathbf{\Sigma} + b^2\mathbf{\Sigma'} \Biggr)
\end{split}
\end{equation}
where $\mathbf{\Sigma'} = \mathbf{I} - c\mathbf{\Sigma}$. The conditional distribution of the latent of next stage starting timestep $\text{Up}(\hat{\mathbf{x}}_{s_{k+1}})$is:
\begin{equation}
\label{append_eq:startpoint_upsample}
\text{Up}(\hat{\mathbf{x}}_{s_{k+1}}) | \mathbf{x}_1 \sim \mathcal{N}\left(s_{k+1}\text{Up}(\mathbf{x}_1), (1-{s_{k+1}})^2 \mathbf{\Sigma}\right)
\end{equation}

Since we want Eq.~\eqref{append_eq:s1} to match Eq.~\eqref{append_eq:startpoint_upsample}, we can obtain the following equations for the mean and standard deviation, respectively:
\begin{equation}
\label{append_eq:mean_eq}
ae_k=s_{k+1},
\end{equation}
\begin{equation}
\label{append_eq:std_eq}
a^2(1-e_k)^2\mathbf{\Sigma}+b^2(\mathbf{I}-c\mathbf{\Sigma})=(1-s_{k+1})^2\mathbf{I}
\end{equation}
From Eqs.~(\ref{append_eq:mean_eq})–(\ref{append_eq:std_eq}), matching the $\mathbf{\Sigma}$ coefficient to zero gives
\begin{equation}
\label{append_eq:A1_1}
a^2(1-e_k)^2 = b^2c
\;\Longrightarrow\;
a(1-e_k)=b\sqrt{c}.
\end{equation}
Comparing the \(\mathbf I\) part,
\begin{equation}
\label{append_eq:A1_2}
b^2=(1-s_{k+1})^2 \;\Longrightarrow\; b=1-s_{k+1}.
\end{equation}
Collecting Eq.~\eqref{append_eq:mean_eq}, Eq.~\eqref{append_eq:A1_1} and Eq.~\eqref{append_eq:A1_2} yields:
\begin{equation}
\label{eq:s}
s_{k+1} = \frac{e_k}{(1-e_k)/\sqrt{c} \; + e_k},
\end{equation}
\begin{equation}
\label{eq:a}
a = \frac{1}{(1-e_k)/\sqrt{c} \; + e_k},
\end{equation}
\begin{equation}
\label{eq:b}
b = \frac{(1-e_k)/\sqrt{c}}{(1-e_k)/\sqrt{c} \; + e_k}.
\end{equation}

By defining the composite term $\delta_k$ as
\begin{equation}
\delta_k \equiv (1-e_k)/\sqrt{c},
\end{equation}
\cref{eq:s},~\cref{eq:a},~\cref{eq:b} can be expressed as:
\begin{equation}
\label{eq:timestep_reschedule_inline}
s_{k+1} = \frac{e_k}{\delta_k + e_k}, \quad a = \frac{1}{\delta_k + e_k}, \quad \text{and} \quad b = \frac{\delta_k}{\delta_k + e_k}.
\end{equation}
Since $\mathbf{\Sigma'} = \mathbf{I} - c\mathbf{\Sigma} \succeq 0$, $0 \leq c \leq 1/4$ for 2× nearest-neighbor upsampling.

Two additional facts can be observed here. 1) If $e_k < 1$, then $s_{k+1} < e_k$. This means that adding noise to the upsampled latent always shifts the timestep towards the noise direction. 2) If $e_k = 1$, then $s_{k+1} = 1$, $a = 1$, and $b = 0$. This implies that upsampling a fully denoised latent results in no subsequent stage. However, in this case, the image quality is reduced because there is no remaining step to refine the error caused by the latent upsample.

\subsection{Values determined by NT-Matching.}
\label{append:values_distribution}
\begin{table}[!b]
    \caption{Determined values of Eq.~\eqref{eq:distribution}.}
    \label{tab:values_distribution}
    \centering
    \resizebox{0.5\textwidth}{!}{
    \begin{scriptsize}
    \begin{tabular}{ccccc}
        \toprule
        \multirow{2}{*}{Model-acceleration} & \multicolumn{2}{c}{values we choose} & \multicolumn{2}{c}{values determined by Eq.~\eqref{eq:distribution}} \\
        \cmidrule(lr){2-5}
        & $N$ & $e$ & $h$ & $c$ \\
        \midrule
        FLUX-5×                & {[}4, 9, 15{]}  & {[}0.3, 0.45, 1.0{]} & {[}5.12, 2.64, 2.25{]}            & 0.0177 \\
        FLUX-7×                & {[}2, 5, 10{]}  & {[}0.2, 0.3, 1.0{]} & {[}8.14, 2.86, 2.19{]}            & 0.0255 \\
        \midrule
        SD3-2×                & {[}5, 11, 20{]}  & {[}0.2, 0.3, 1.0{]} & {[}6.21, 2.23, 1.97{]}            & 0.0586 \\
        SD3-3×                & {[}3, 6, 12{]}  & {[}0.25, 0.3, 1.0{]} & {[}6.40, 2.60, 2.23{]}            & 0.0255 \\
        \bottomrule
    \end{tabular}
    \end{scriptsize}
    }
\end{table}

\begin{equation}
\{h_k\}, c = \argmin_{\{h_k\}, c} \mathrm{JSD}(P_{target}(t),P(t)).
\label{eq:distribution}
\end{equation}
We determine the values of \{$h_k$\} and 
$c$ in Sec.~4.2 by minimizing the Jensen-Shannon divergence (JSD) between $P_{target}(t)$ (Eq.~(11)) and $P(t)$ (Eq.~(12)). The resulting values are showed in Tab.~\ref{tab:values_distribution}.

\section{Detailed Experimental Setup}
\subsection{Experiment compute resources}
\label{append:resources}
We generate 1024 $\times$ 1024 images using an NVIDIA A100 GPU for all experiments. The latency results reported in Tab.~1 are benchmarked on a single A100 GPU.
Furthermore, to demonstrate that RALU is generalizable across different hardware architectures, we also benchmark latency on an NVIDIA A6000 GPU in~\cref{tab:compute}. In addition, we report GPU memory usage.

\begin{table}[!h]
    \centering
    \caption{Computational metrics (FLOPs, GPU peak memory) with mixed precision FP16. FLOPs are measured using the \texttt{torch.profiler} tool. GPU peak memory is measured using NVIDIA A6000 GPU. We employed FlashAttention~\cite{dao2022flashattention} for all models excluding ToCa~\cite{zou2024accelerating}.}
    \label{tab:compute}
    \resizebox{0.49\textwidth}{!}{
    \begin{scriptsize}
    \begin{tabular}{ccccc}
         \toprule
        \multirow{2}{*}{Method} & \multirow{2}{*}{Accel.} & \multicolumn{2}{c}{Latency (s)} &  \multirow{2}{*}{VRAM usage (GB)} \\
        \cmidrule(lr){3-4}
        & & A100 & A6000  \\
        \midrule
        FLUX (50) & - & 25.1 & 49.7 & 32.8 \\
        \cmidrule(lr){1-5}
        FLUX (10) & \textcolor{blue}{\textbf{T}} & 5.18 & 9.99 & 32.8 \\
        $\Delta$-DiT~\cite{deltadit} & \textcolor{blue}{\textbf{T}} & 7.42 & 10.6 & 45.0 \\
        ToCa~\cite{zou2024accelerating} & \textcolor{blue}{\textbf{T}} & 15.5 & 29.7 & 45.0 \\
        TeaCache~\cite{liu2025timestep} & \textcolor{blue}{\textbf{T}} & 5.23 & 9.91 & 34.8 \\
        TaylorSeer~\cite{liu2025reusing} & \textcolor{blue}{\textbf{T}} & 9.34 & 17.3 & 45.0 \\
        Bottleneck~\cite{tian2025training} & \textcolor{red}{\textbf{S}} & 5.37 & 10.3 & 32.8 \\
        \textbf{RALU (Ours)} & \textcolor{red}{\textbf{S}} & 5.04 & 9.60 & 32.8 \\
        \cmidrule(lr){1-5}
        FLUX (7) & \textcolor{blue}{\textbf{T}} & 3.79 & 7.21 & 32.8 \\
        TeaCache~\cite{liu2025timestep}  & \textcolor{blue}{\textbf{T}} & 4.21 & 7.90 & 34.8 \\
        TaylorSeer~\cite{liu2025reusing} & \textcolor{blue}{\textbf{T}} & 7.00 & 12.4 & 45.0 \\
        Bottleneck~\cite{tian2025training} & \textcolor{red}{\textbf{S}} & 3.78 & 7.21 & 32.8 \\
        \textbf{RALU (Ours)} & \textcolor{red}{\textbf{S}} & 3.75 & 7.16 & 32.8\\
        \bottomrule
    \end{tabular}
    \end{scriptsize}
    }
    \resizebox{0.49\textwidth}{!}{
    \begin{scriptsize}
    \begin{tabular}{ccccc}
         \toprule
        \multirow{2}{*}{Method} & \multirow{2}{*}{Accel.} & \multicolumn{2}{c}{Latency (s)} &  \multirow{2}{*}{VRAM usage (GB)} \\
        \cmidrule(lr){3-4}
        & & A100 & A6000  \\
        \midrule
        SD3 (28) & - & 4.04 & 6.77 & 15.5 \\
        \cmidrule(lr){1-5}
        SD3 (14) & \textcolor{blue}{\textbf{T}} & 2.14 & 3.57 & 15.5 \\
        $\Delta$-DiT~\cite{deltadit}  & \textcolor{blue}{\textbf{T}} & 2.41 & 4.04 & 16.7 \\
        ToCa~\cite{zou2024accelerating} & \textcolor{blue}{\textbf{T}} & 2.62 & 4.32 & 17.6 \\
        RAS~\cite{liu2025region}  & \textcolor{blue}{\textbf{T}} & 2.07 & 3.45 & 20.7 \\
        TaylorSeer~\cite{liu2025reusing} & \textcolor{blue}{\textbf{T}} & 2.40 & 4.00 & 17.9 \\
        Bottleneck~\cite{tian2025training} & \textcolor{red}{\textbf{S}} & 2.41 & 3.57 & 15.5 \\
        \textbf{RALU (Ours)} & \textcolor{red}{\textbf{S}} & 2.04 & 3.41 & 15.5  \\
        \cmidrule(lr){1-5}
        SD3 (9) & \textcolor{blue}{\textbf{T}} & 1.46 & 2.39 & 15.5 \\
        RAS~\cite{liu2025region}  & \textcolor{blue}{\textbf{T}} & 1.40 & 2.33 & 20.7 \\
        TaylorSeer~\cite{liu2025reusing} & \textcolor{blue}{\textbf{T}} & 1.87 & 3.11 & 17.9 \\
        Bottleneck~\cite{tian2025training} & \textcolor{red}{\textbf{S}} & 1.44 & 2.40 & 15.5 \\
        \textbf{RALU (Ours)} & \textcolor{red}{\textbf{S}} & 1.38 & 2.24 & 15.5 \\
        \bottomrule
    \end{tabular}
    \end{scriptsize}
    }
\end{table}

\subsection{Baseline configurations}
\subsubsection{Main experiments (Tab.1, 2)}
\label{append:baselines_main}

\paragraph{$\Delta$-DiT~\cite{deltadit}}
$\Delta$-DiT is a training-free inference acceleration framework specifically designed for the Diffusion Transformers (DiT) architecture. To address the information loss issues of prior U-Net-based caching, $\Delta$-DiT introduces the $\Delta$-Cache mechanism, which is suitable for DiT's isotropic structure. $\Delta$-Cache selectively caches the deviation (offset) between feature maps instead of the feature maps themselves, thus preserving critical information from the previous sampling step. The core strategy is Stage-Adaptive Acceleration, based on the finding that DiT's front blocks are associated with generating image outlines while the rear blocks handle details. This knowledge is aligned with the diffusion process: $\Delta$-Cache is applied to the rear blocks during the early sampling stages (outline-friendly), and to the front blocks during the later sampling stages (detail-friendly). We set $N=5$ for FLUX 5$\times$ acceleration and $N=2$ for SD3 2$\times$ acceleration, where $N$ is the interval between fully computed steps.

\paragraph{ToCa~\cite{zou2024accelerating}}
Token-wise feature Caching (ToCa) is a training-free inference-time acceleration method for Diffusion Transformers that improves efficiency through token-level feature caching. Unlike na\"ive caching methods that reuse all token features uniformly across timesteps, ToCa selectively caches tokens based on their importance, which is determined by their influence on other tokens (via self-attention), their association with conditioning signals (via cross-attention), their recent cache frequency, and their spatial distribution within the image.
ToCa operates by dividing inference into cache periods of length $N$, where full computation is performed at the first step and cached token features are reused for the next $N-1$ steps. Within each timestep, a fraction $R$ of the tokens---those deemed less important based on self-attention, cross-attention, cache frequency, and spatial distribution---are selected for caching, while the remaining tokens are recomputed.
We set $N=5$, $N_\text{total}=40$, $R=90\%$ for FLUX 5$\times$ acceleration and $N=3$, $N_\text{total}=28$, $R=90\%$ for SD3 2$\times$ acceleration, where $N_\text{total}$ is total inference steps.

\paragraph{TeaCache~\cite{liu2025timestep}}
TeaCache (Timestep Embedding Aware Cache) is a training-free caching approach designed to accelerate diffusion models by selectively reusing intermediate model outputs. It addresses the inflexibility of conventional uniform caching by leveraging the fact that model output differences fluctuate non-uniformly across timesteps. The core idea is to predict output difference using model inputs, which have negligible computational cost. TeaCache uses the timestep-embedding modulated noisy input as the primary indicator for output caching, as this composite input shows a strong correlation with the model output difference. To correct a scaling bias between the estimated input difference and the true output difference, TeaCache employs a simple polynomial fitting procedure. Caching is determined dynamically using an accumulated relative $\text{L1}$ distance against a threshold ($\delta$), enabling the skipping of redundant computations in consecutive timesteps. We set $N_\text{total}=45$, $\delta=0.99$ for FLUX 5$\times$ acceleration and $N_\text{total}=30$, $\delta=0.99$ for FLUX 7$\times$ acceleration, where $N_\text{total}$ is total inference steps.

\paragraph{RAS~\cite{liu2025region}}
Region-Adaptive Sampling (RAS) is a training-free inference-time acceleration method for Diffusion Transformers that dynamically adjusts the sampling ratio for different spatial regions. At each diffusion step, RAS identifies fast-update regions—typically semantically important areas—based on the model’s output noise and attention continuity across steps. These regions are refined using the DiT model, while slow-update regions reuse cached noise from the previous step to save computation.
To prevent error accumulation in ignored regions, RAS periodically resets all regions through dense steps. Additionally, RAS employs dynamic sampling schedules (e.g., full updates in early steps and gradual reduction thereafter) and key-value caching in attention to maintain quality.
RAS dynamically determines which spatial regions require refinement at each step by identifying fast-update areas based on noise deviation and attention continuity. The sampling ratio denotes the proportion of tokens actively updated by the DiT model in each step, while the remaining tokens reuse previously cached noise to reduce computation. We set the sampling ratio to $0.32$ for SD3 2$\times$ acceleration and $0.05$ for SD3 3$\times$ acceleration.

\paragraph{TaylorSeer~\cite{liu2025reusing}}
Taylorseer introduces a new "cache-then-forecast" paradigm to accelerate DiTs, overcoming the severe quality degradation of prior "cache-then-reuse" methods at high acceleration ratios. It is a training-free approach. The core idea is based on the observation that features in diffusion models evolve along a stable and continuous trajectory across timesteps.
Taylorseer leverages Taylor series expansion to predict the features at future timesteps based on cached values from previous steps. The parameter $N$ is the interval (or forced activation period) between fully computed steps. Furthermore, by using an order $O$ greater than 0, the method uses finite difference methods to approximate the features' higher-order derivatives. This enables more accurate modeling of the nonlinear feature trajectory and reduces cumulative prediction errors, which is especially crucial over large intervals (a large $N$). This predictive strategy allows the model to maintain generation quality even at extreme speedups. We set $N=7$, $O=1$, $W=3$ for FLUX 5$\times$ acceleration, $N=11$, $O=1$, $W=3$ for FLUX 7$\times$ acceleration, $N=2$, $O=1$, $W=3$ for SD3 2$\times$ acceleration, and $N=4$, $O=1$, $W=3$ for SD3 3$\times$ acceleration, where $W$ is the number of fully warm-up steps.

\paragraph{Bottleneck Sampling~\cite{tian2025training}}
Bottleneck Sampling is a training-free, inference-time acceleration method that exploits the low-resolution priors of pre-trained diffusion models. It adopts a three-stage high–low–high resolution strategy: starting with high-resolution denoising to establish semantic structure, performing low-resolution denoising in the intermediate steps to reduce computational cost, and restoring full resolution at the final stage to refine details.
To ensure stable denoising across stage transitions, Bottleneck Sampling introduces two key techniques: (1) noise reintroduction, which resets the signal-to-noise ratio (SNR) at each resolution change to avoid inconsistencies, and (2) scheduler re-shifting, which adapts the denoising schedule per stage to align with the changed resolution and noise levels.
We set the cumulative number of inference steps at the end of each stage $[N_1, N_2, N_3] = [4, 16, 21]$ for FLUX 5$\times$ acceleration, $[3, 11, 15]$ for FLUX 7$\times$ acceleration, $[6, 24, 31]$ for SD3 2$\times$ acceleration, and $[4, 16, 20]$ for SD3 3$\times$ acceleration.


\subsubsection{Combining RALU with other accelerations (Tab.3)}
\label{append:baselines_combining}
To evaluate the feasibility and performance of integrating RALU with various acceleration methods, we established several baselines.

Our primary focus was on combining RALU with temporal acceleration methods, for which we employed $\Delta$-DiT, ToCa, and TaylorSeer. The specific configurations for these baselines were as follows:
\begin{itemize}
\item
$\Delta$-DiT: The caching interval, denoted as $N$, was set to $N=2$.
\item
ToCa: We configured the caching interval to $N=3$ and the skip ratio, $R$ (representing the percentage of frames to skip), to $R=90\%$.
\item
TaylorSeer: The forced activation period was set to $N=3$, and the order of the Taylor expansion, $O$, was set to $O=1$.
\end{itemize}

We additionally integrated our method with timestep-distilled models by employing the publicly available weights of FLUX.1-schnell and SD3.5L-Turbo from Hugging Face. For both models, we set the number of function evaluations (NFE) to 4 and configured the cumulative number of inference steps at the end of each stage as $[N_1, N_2, N_3] = [1, 2, 4]$ and $e = [0.3, 0.45, 1.0]$.

\subsection{Flow-matching based Diffusion Transformers}
For the quantitative comparison, we performed experiments on two flow-matching based diffusion transformers (DiTs)~\cite{peebles2023scalable}: FLUX-1.dev~\cite{black2024flux} and Stable Diffusion 3~\cite{peebles2023scalable}.

\paragraph{FLUX.1-dev}
FLUX.1-dev is a diffusion-based text-to-image (T2I) synthesis model trained on large-scale data via flow matching, achieving state-of-the-art performance. Despite its high generation quality, the model combines T5-XXL~\cite{raffel2020exploring} and a CLIP~\cite{radford2021learning} text encoder, resulting in a total of 12 billion parameters. This large model size leads to significant inference latency, posing serious limitations for real-world deployment. In this work, we apply various acceleration methods, including our proposed approach, to FLUX.1-dev and evaluate each method in terms of image quality and faithfulness to the input text. These evaluations demonstrate the effectiveness of our method.

\paragraph{Stable Diffusion 3}
Stable Diffusion 3 (SD3) is a text-to-image synthesis diffusion generative model trained with a rectified flow objective. It conditions on three different text encoders—CLIP-L~\cite{radford2021learning}, CLIP-G, and T5-XXL~\cite{raffel2020exploring}—and has a total of 8 billion parameters. Due to its large model size, SD3 also suffers from non-negligible inference latency, which remains one of the key challenges. In this work, we conduct experiments on SD3 with 2× and 3× speedups to evaluate the effectiveness of our proposed method. In our experiments, we use Stable Diffusion 3 Medium.

\subsection{Metrics}
\label{append:metrics}



\paragraph{ImageReward~\cite{xu2023imagereward}}
ImageReward is presented as the first general-purpose text-to-image human preference reward model (RM), designed to effectively learn and encode human preferences for evaluating and improving text-to-image generation models. Its training relies on a systematic annotation pipeline, including rating and ranking, which collected 137k expert comparisons based on real-world user prompts. ImageReward comprehensively captures human preference by evaluating multiple factors, including text-image alignment, image quality, and Harmlessness. ImageReward serves two key roles: as a promising automatic evaluation metric for comparing text-to-image models and individual samples, and as a reward function to directly optimize diffusion models via the Reward Feedback Learning (ReFL) algorithm.
In our experiments, we average the ImageReward scores over 5,000 images using MS-COCO validation set~\cite{lin2014microsoft}.


\begin{table*}[t!]
    \centering
    \caption{Comparison of different edge detection methods on FLUX.1-dev.}
    \label{tab:edge_detection_methods}
    \resizebox{0.95\textwidth}{!}{
    \begin{scriptsize}
    \begin{tabular}{c|cccccc}
        \toprule
        Edge detection methods & TFLOPs $\downarrow$ & ImageReward $\uparrow$ & CLIP-IQA $\uparrow$ & NIQE $\downarrow$ & T2I-Comp. $\uparrow$ & GenEval $\uparrow$\\
        \midrule
        gradient kernel (5$\times$) & \textbf{537.99} & 0.996 & 0.694 & 6.77 & \textbf{0.627} & \textbf{0.675} \\
        \textbf{VAE decode $\xrightarrow{}$ Canny} (5$\times$) & 540.47 & \textbf{1.022} & \textbf{0.700} & \textbf{6.43} & 0.626 & 0.652 \\
        \midrule
        gradient kernel (7$\times$) & \textbf{423.53} & 0.979 & \textbf{0.685} & 6.98 & \textbf{0.633} & 0.662 \\
        \textbf{VAE decode $\xrightarrow{}$ Canny} (7$\times$) & 426.01 & \textbf{0.999} & 0.681 & \textbf{6.87} & \textbf{0.633} & \textbf{0.682} \\
        \bottomrule
    \end{tabular}
    \end{scriptsize}
    }
\end{table*}

\begin{table*}[t!]
    \centering
    \caption{Comparison of method to determine early upsampling patches on FLUX.1-dev. FLOPs are reported as Mean $\pm$ Std.}
    \label{tab:adaptive_upsample}
    \resizebox{0.95\textwidth}{!}{
    \begin{scriptsize}
    \begin{tabular}{c|ccccccc}
        \toprule
        Early upsampling ratio & TFLOPs $\downarrow$ & ImageReward $\uparrow$ & CLIP-IQA $\uparrow$ & NIQE $\downarrow$ & T2I-Comp. $\uparrow$ & GenEval $\uparrow$\\
        \midrule
        Adaptive ratio (5$\times$) & \tightcell{550.92}{22.91} & 1.015 & 0.691 & 6.56 & \textbf{0.630} & \textbf{0.668} \\
        \textbf{Fixed} $\textbf{r=0.2}$ (5$\times$) & \textbf{\tightcell{540.47}{0.00}} & \textbf{1.022} & \textbf{0.700} & \textbf{6.43} & 0.626 & 0.652 \\
        \midrule
         Adaptive ratio (7$\times$) & \tightcell{435.14}{12.36} & 0.986 & 0.677 & \textbf{6.81} & \textbf{0.634} & 0.674 \\
        \textbf{Fixed} $\textbf{r=0.2}$ (7$\times$) & \textbf{\tightcell{426.01}{0.00}} & \textbf{0.999} & \textbf{0.681} & 6.87 & 0.633 & \textbf{0.682} \\
        \bottomrule
    \end{tabular}
    \end{scriptsize}
    }
\end{table*}

\paragraph{GenEval~\cite{ghosh2023geneval}}
GenEval is a comprehensive benchmark designed to evaluate the alignment between generated images and input text prompts in text-to-image (T2I) synthesis. We use two-object, counting, color, and color attribution prompts to evaluate models. It is designed to specifically probe the compositional understanding of T2I models by leveraging existing object detection and other discriminative vision models to verify properties. It can assess how faithfully the generated outputs reflect the semantic content of the given textual descriptions.
In our experiments, each prompt is sampled with four different random seeds.









\paragraph{T2I-CompBench~\cite{huang2023t2i}}
T2I-CompBench is a benchmark specifically designed to assess the compositional understanding of T2I generation models. It comprises structured prompts aimed at evaluating a model’s ability to accurately associate attributes with corresponding objects, ensuring correct semantic alignment in scenarios involving multiple objects and attributes.
For evaluation, we measured performance on spatial, non-spatial and complex sets.
By presenting diverse and challenging prompts, T2I-CompBench offers a rigorous evaluation framework for diagnosing issues such as semantic neglect that are prevalent in T2I models.
For quantitative evaluation, each prompt is sampled with four different random seeds.


\paragraph{CLIP-IQA~\cite{wang2023exploring}}
The CLIP-IQA metric leverages the pre-trained vision-language model CLIP to assess both quality and abstract perception of images without task-specific training. By using a novel antonym prompt pairing strategy (\textit{e.g.}, ``Good photo.'' vs. ``Bad photo.'') and removing positional embeddings to accommodate variable input sizes, CLIP-IQA computes the perceptual similarity between images and descriptive prompts. This enables it to evaluate traditional quality attributes like sharpness and brightness as well as abstract attributes such as aesthetic or emotional tone. Extensive evaluations on standard IQA benchmarks and user studies suggest that CLIP-IQA achieves competitive correlation with human perception compared to established no-reference and learning-based methods, while maintaining generality and flexibility. We utilize CLIP-IQA provided by PyIQA~\footnote{\label{pyiqa}\url{https://github.com/chaofengc/IQA-PyTorch}} with the default prompt setting.
We average the CLIP-IQA scores over 5,000 generated images using MS-COCO validation set~\cite{lin2014microsoft}.

\paragraph{NIQE~\cite{mittal2012making}}
The Natural Image Quality Evaluator (NIQE) is a no-reference image quality assessment (IQA) metric that operates without any training on human opinion scores or exposure to distorted images. It is a completely blind, opinion-unaware, and distortion-unaware model that measures deviations from statistical regularities observed in natural images. NIQE extracts perceptually relevant natural scene statistics (NSS) features from local image patches and fits them to a multivariate Gaussian (MVG) model built from a corpus of pristine images. The image quality is then quantified as the distance between the MVG model of the test image and that of natural images. Unlike many existing no reference IQA models that are limited to distortion types seen during training, NIQE is general-purpose and performs competitively with state-of-the-art methods, while requiring no supervised learning and maintaining low computational complexity.
We utilize NIQE provided by PyIQA~\footref{pyiqa} with the default prompt setting.
We average the NIQE scores over 5,000 images using MS-COCO validation set~\cite{lin2014microsoft}.

\section{Additional Experiments}




\subsection{Additional ablation studies}

\paragraph{Effect of edge detection method.}
To mitigate aliasing artifacts, we adopt an early upsampling strategy focused on artifact-prone edge regions. In this process, we initially approximate the clean latent ($\hat{x_0}$) using Tweedie's formula, which is then decoded by the VAE to apply Canny edge detection in the image space. A natural question arises: \textit{Is VAE decoding strictly necessary?}

Detecting edges directly in the latent space without VAE decoding is inherently inaccurate. High-frequency content in the latent space does not consistently correspond to high-frequency edges in the image space, given the implicit and compressed nature of the information. However, an alternative approach is to utilize a gradient kernel to measure the steepness of change (gradient) between adjacent latent patches and then hypothesize that regions exhibiting a large magnitude of change constitute the edge regions.

Tab.~\ref{tab:edge_detection_methods} compares FLOPs and performance metrics of different edge detection methods. The gradient kernel method is marginally more efficient in terms of FLOPs as it avoids the VAE decode step.
Nevertheless, we observe that the VAE decoding followed by Canny edge detection yields better performance across overall metrics.
Despite the slightly higher computational overhead, this method adds a minimal amount of computation (i.e., 2.48 TFLOPs), accounting for less than $1\%$ of the total FLOPs. Therefore, we adopt the VAE decoding and Canny edge detection approach for its higher accuracy in identifying edge regions.

\begin{table}[t]
    \centering
    \caption{Comparison of different upsampling methods on FLUX.1-dev, two-stage upsampling framework. We denote the best performance in \textbf{bold} and second-best performance with \underline{underline}.}
    \label{tab:upsampling_methods}
    \resizebox{\linewidth}{!}{
    \begin{footnotesize}
    \begin{tabular}{c|cc}
        \toprule
        Upsampling methods & ImageReward $\uparrow$ & CLIP-IQA $\uparrow$ \\
        \midrule
        Bilinear & \underline{0.9883} & 0.7045 \\
        Bicubic & 0.9754 & 0.7032\\
        Lanczos & 0.9848 & \textbf{0.7077} \\
        \textbf{Nearest-neighbor (Ours)} & \textbf{0.9916} & \underline{0.7056} \\
        \bottomrule
    \end{tabular}
    \end{footnotesize}
    }
\end{table}

\paragraph{Adaptive upsampling ratio.}
We select fixed upsampling ratio top-$r$, where $r=0.2$ for FLUX-5$\times$, FLUX-7$\times$, SD3-3$\times$ settings and $r=0.3$ for SD3-2$\times$ setting. However, recognizing that the proportion of edge patches can vary across images, we also experimented with an adaptive selection strategy based on a fixed edge-strength threshold, rather than a fixed top-r ratio, to determine if this improved image quality.
The results in~\cref{tab:adaptive_upsample} show that this adaptive approach did not yield improvements in image fidelity although this method introduces computational instability (i.e., high variance in FLOPs) with a slightly higher average computational cost.
Given that the fixed-ratio method ensures consistent, predictable computational costs and delivers superior or comparable generation quality, we adopted it for its stability and more favorable performance-efficiency trade-off.

\paragraph{Effect of upsampling method.}
We employed nearest-neighbor upsampling for feature upsampling. This choice is motivated by the inherent risks associated with conventional image-space interpolation methods when operating in the latent space, where feature correlations are highly complex.

We conducted comparisons with various upsampling methods. Specifically, NT-Matching involves injecting correlated noise, necessitate calculating the exact covariance matrix for the noise. However, this calculation introduces prohibitive computational overhead, rendering it unsuitable for DiT acceleration. Consequently, for a fair comparison in this context, we focused on matching the Signal-to-Noise Ratio (SNR).
Tab.~\ref{tab:upsampling_methods} shows that employing nearest-neighbor upsampling yields better performance than other upsamplers. Furthermore, the nearest-neighbor approach simplifies the computation of the covariance matrix, which ultimately led to our choice of this upsampling method.

\begin{figure}[t]
    \centering
    \begin{tikzpicture}[
        promptblock/.style={font=\footnotesize, align=center, text width=7.5cm},
        modelblock/.style={font=\footnotesize, align=center, text width=1cm}]
        \node[inner sep=0pt] (p1a) at (0, 0) {\includegraphics[width=0.31\linewidth]{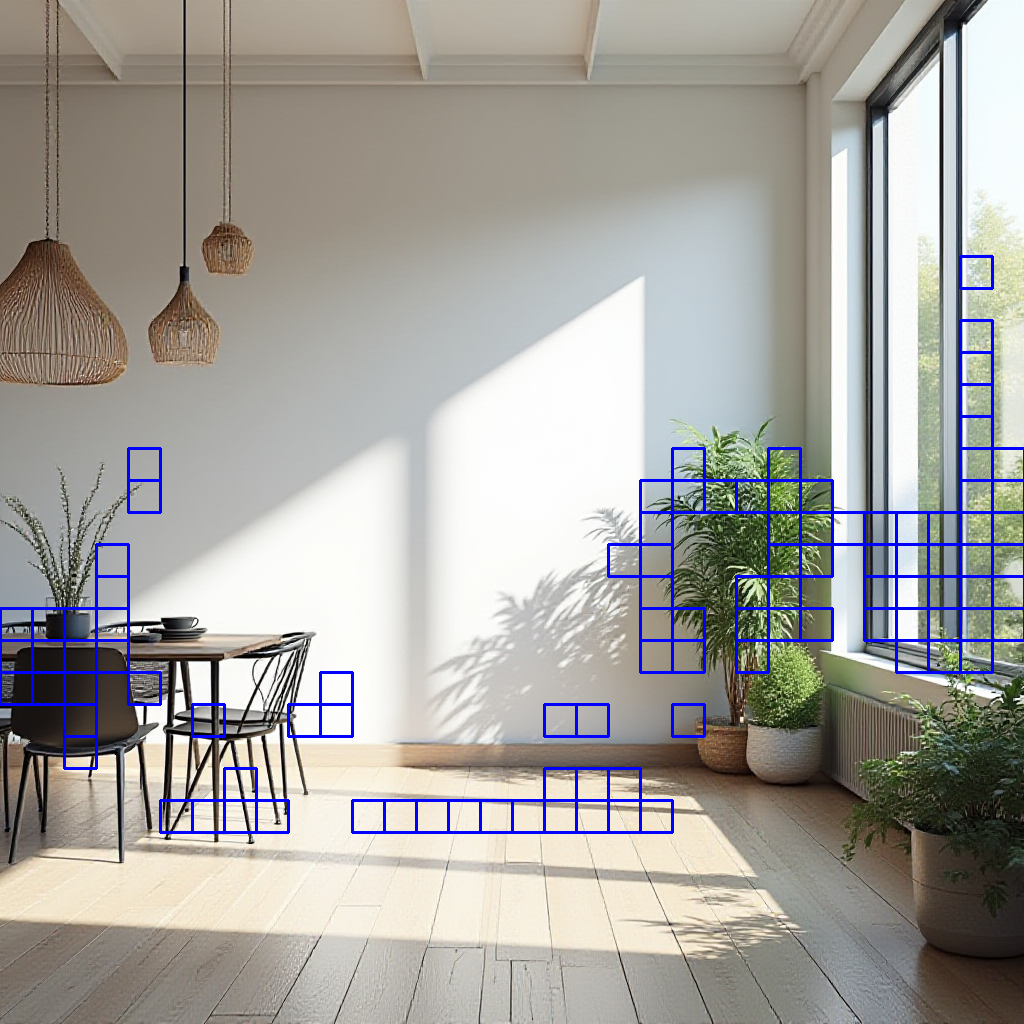}};
        \node[inner sep=0pt] (p2a) at (2.9, 0) {\includegraphics[width=0.31\linewidth]{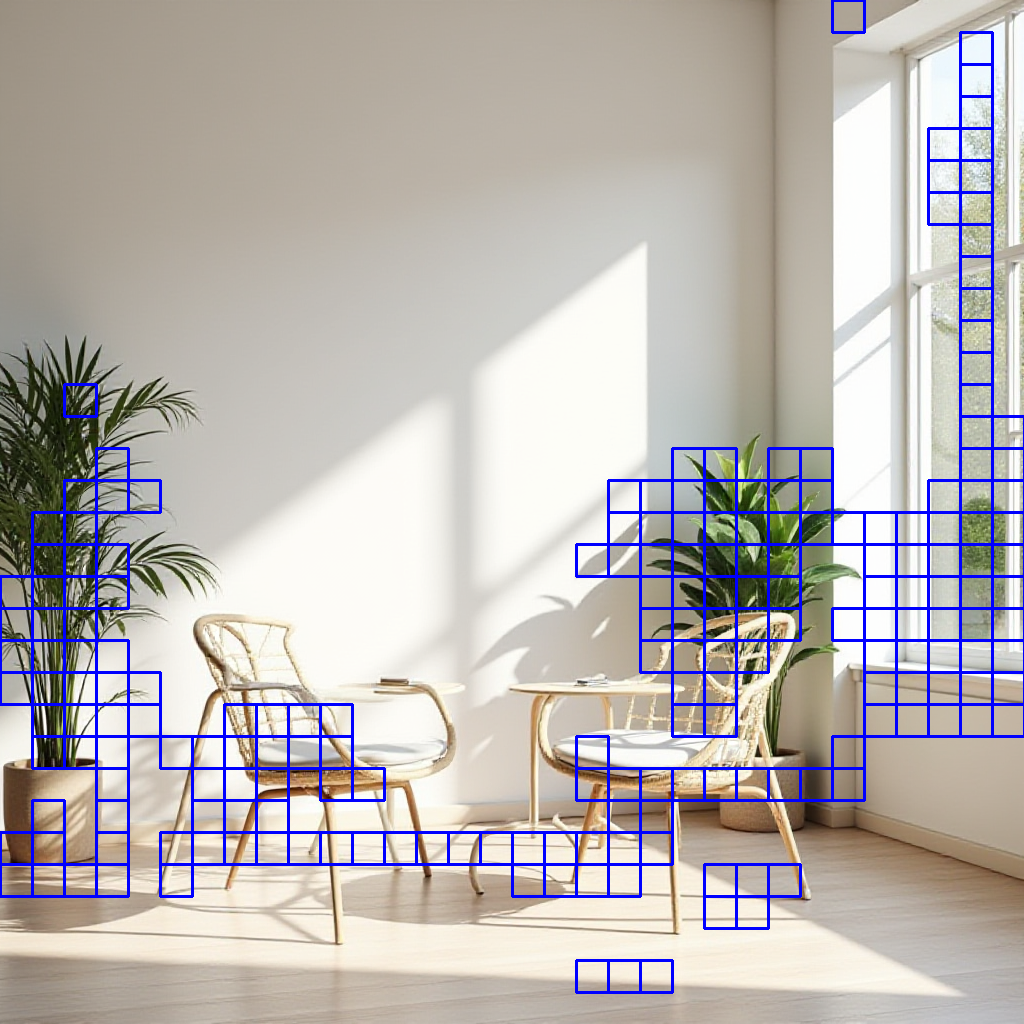}};
        \node[inner sep=0pt] (p3a) at (5.8, 0) {\includegraphics[width=0.31\linewidth]{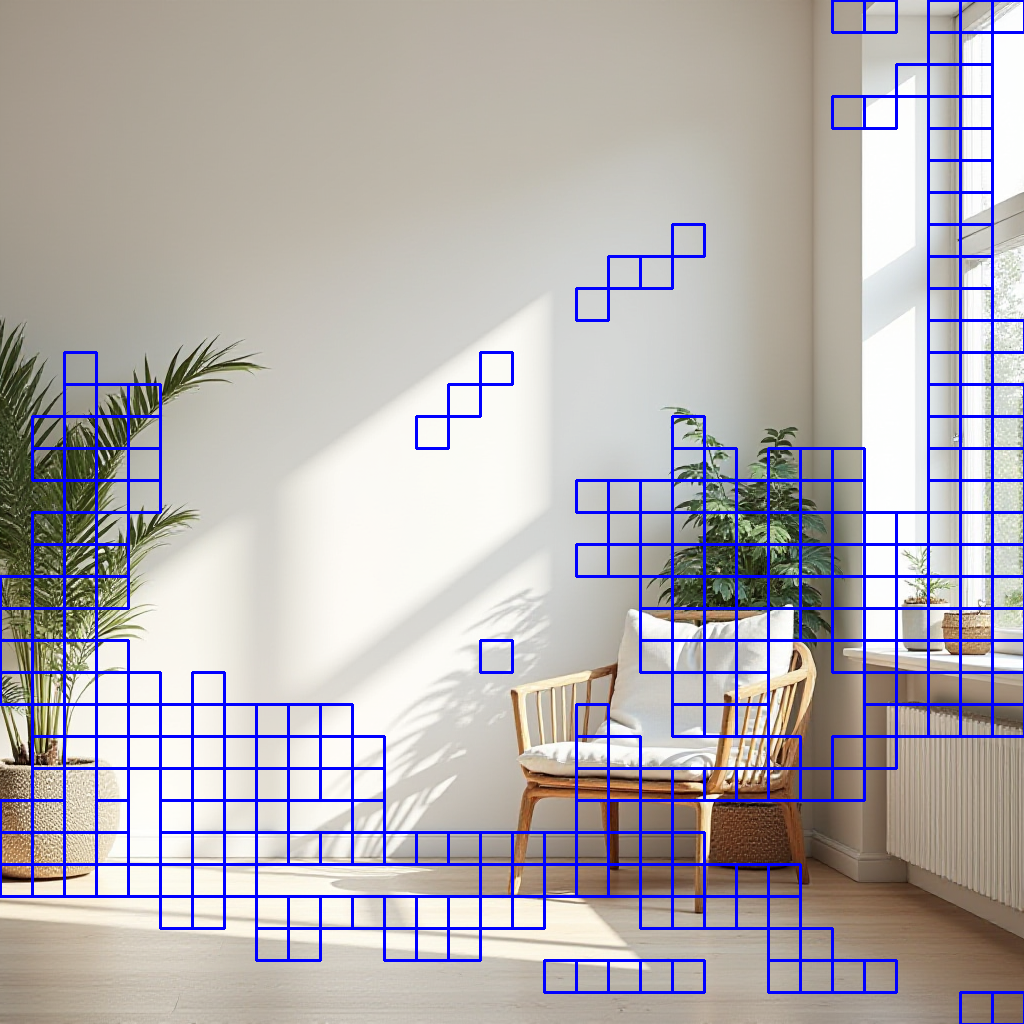}};
        \node[inner sep=0pt] (p1b) at (0, -3.5) {\includegraphics[width=0.31\linewidth]{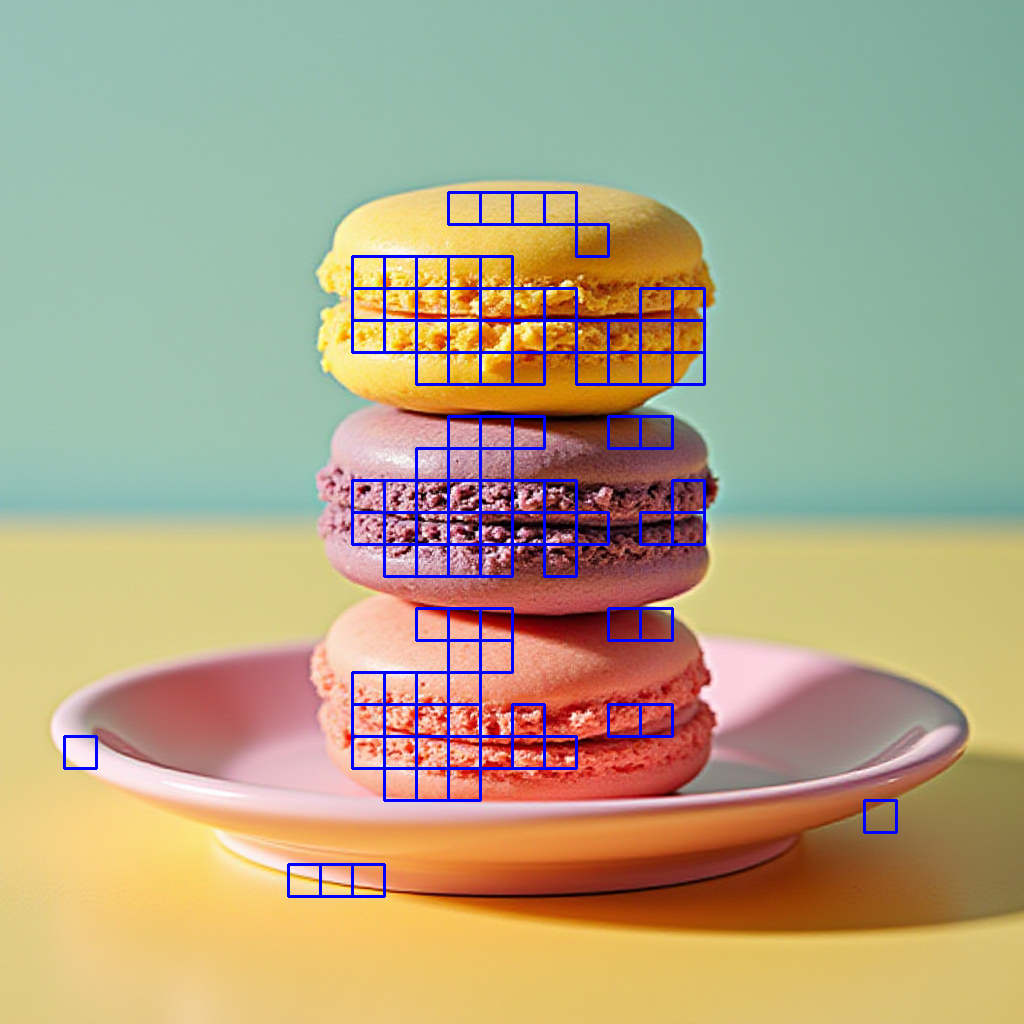}};
        \node[inner sep=0pt] (p2b) at (2.9, -3.5) {\includegraphics[width=0.31\linewidth]{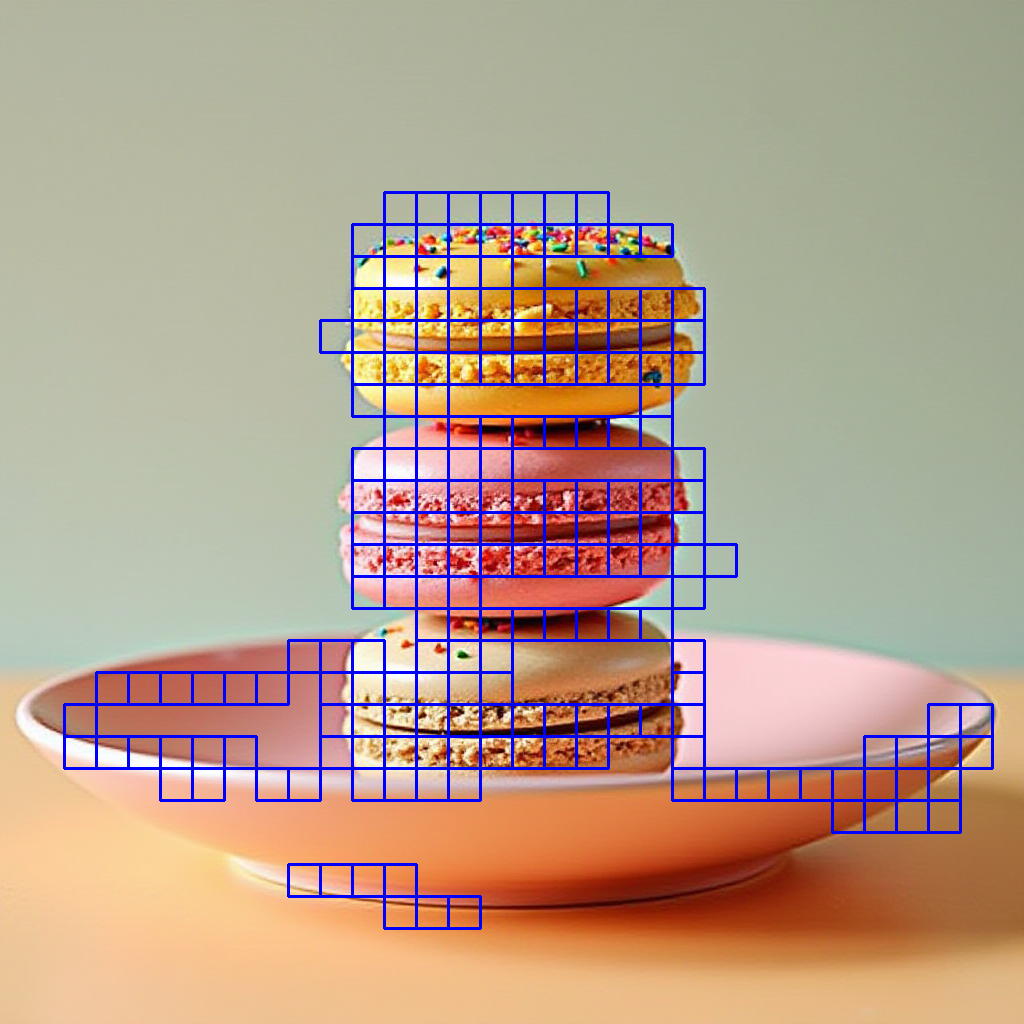}};
        \node[inner sep=0pt] (p3b) at (5.8, -3.5) {\includegraphics[width=0.31\linewidth]{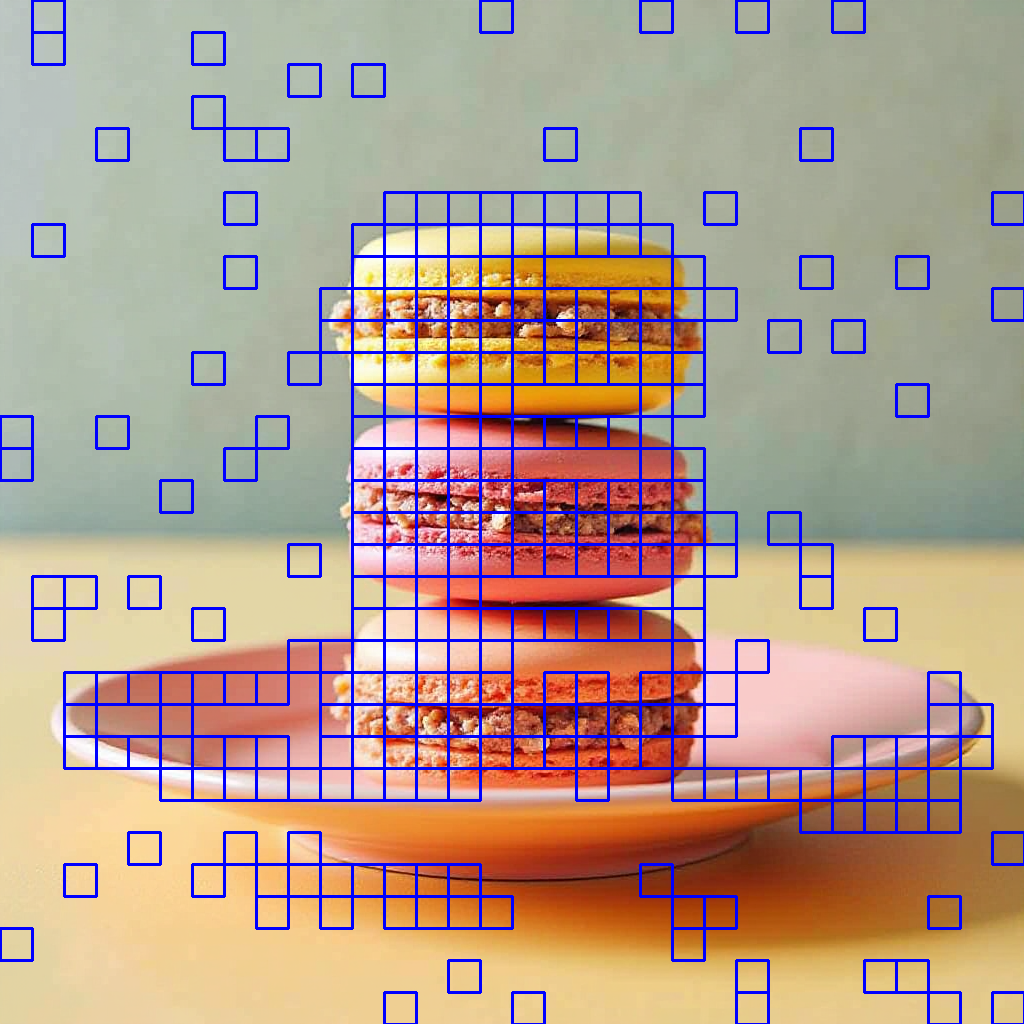}};
        
        \node[promptblock] at (2.9, -1.65)
        {\emph{``A bright, sunlit room with a few pieces of modern furniture.''}};
        \node[promptblock] at (2.9, -5.15)
        {\emph{``Three colorful macaroons stacked neatly on a bright plate.''}};

        \node[modelblock] at (0, 1.5)
        {$r=0.1$};
        \node[modelblock] at (2.9, 1.5)
        {$r=0.2$};
        \node[modelblock] at (5.8, 1.5)
        {$r=0.3$};
    \end{tikzpicture}
    \caption{Visualization of detected edge regions under different top-$r$ ratios. The patches identified by the Canny edge detector are highlighted with blue squares. We observe that at $r=0.3$, the detector effectively captures the majority of structural edges, while also including some non-edge regions. This visualization is conducted under the FLUX-5$\times$ acceleration setting.}
    \label{fig:edge_visualize}
\end{figure}

\subsection{Visualization of edge detection}
As discussed in~\cref{tab:edge_detection_methods}, applying Canny edge detection after VAE decoding effectively identifies the edge regions. To better understand which areas are actually detected as edge regions, we visualize the detection results in~\cref{fig:edge_visualize}. The figure illustrates the detected edge regions under different early upsampling ratios $r$.

\subsection{Integrating RALU with temporal acceleration}
\begin{figure*}[!t]
  \centering
  \includegraphics[width=1.0\linewidth]{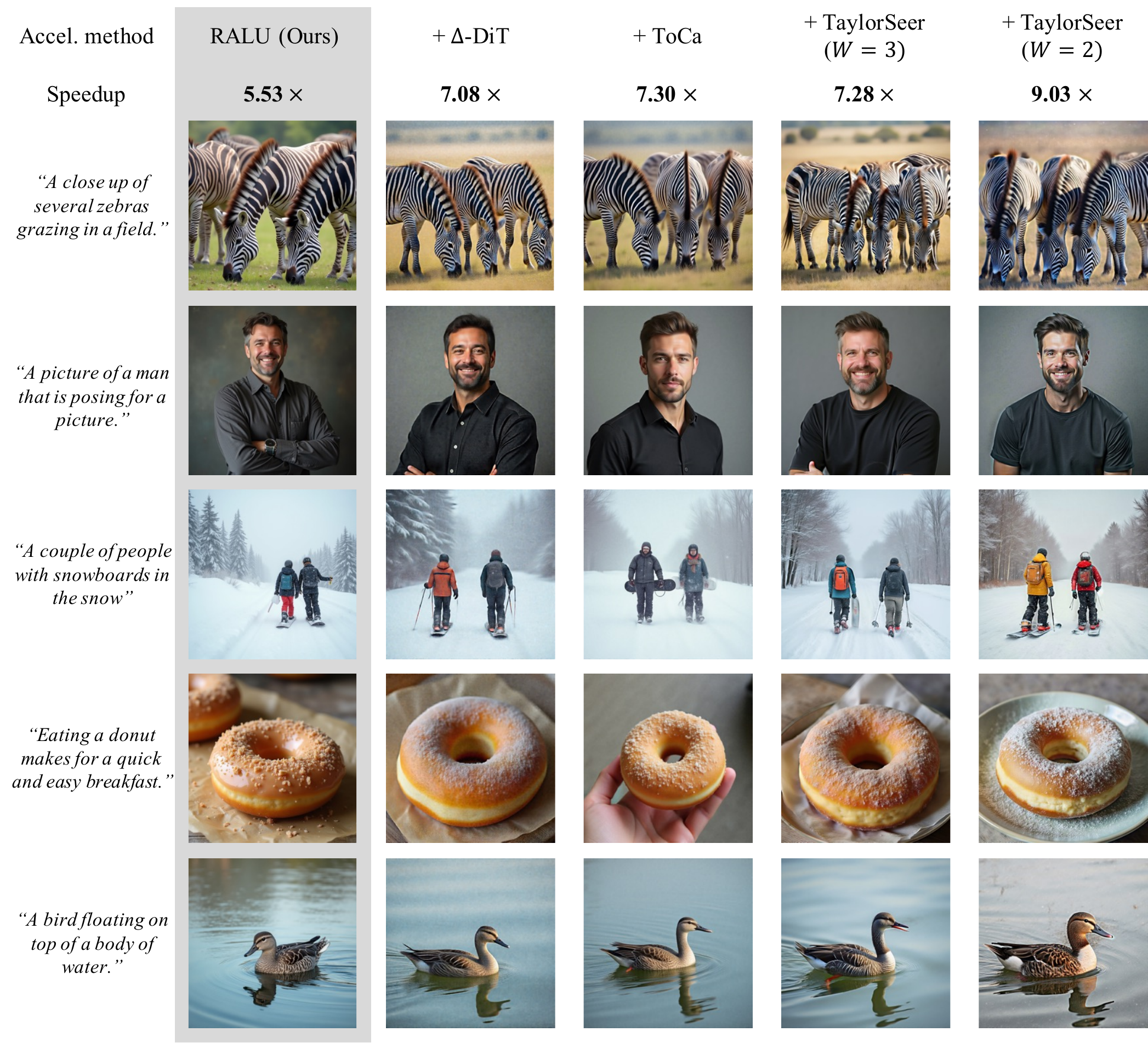}
    \caption{Qualitative comparison of RALU integrated with different temporal acceleration methods.}
    \label{fig:supp_temporal_quali}
  \hfill
\end{figure*}

Fig.~\ref{fig:supp_temporal_quali} presents qualitative comparisons obtained by integrating RALU with several temporal acceleration methods. Although the overall speedup increases from 5.53$\times$ to as high as 9.03$\times$, the image quality remains well preserved, and no noticeable artifacts are introduced.

\subsection{Integrating RALU with timestep-distilled models}
\begin{figure*}[!t]
  \centering
  \includegraphics[width=1.0\linewidth]{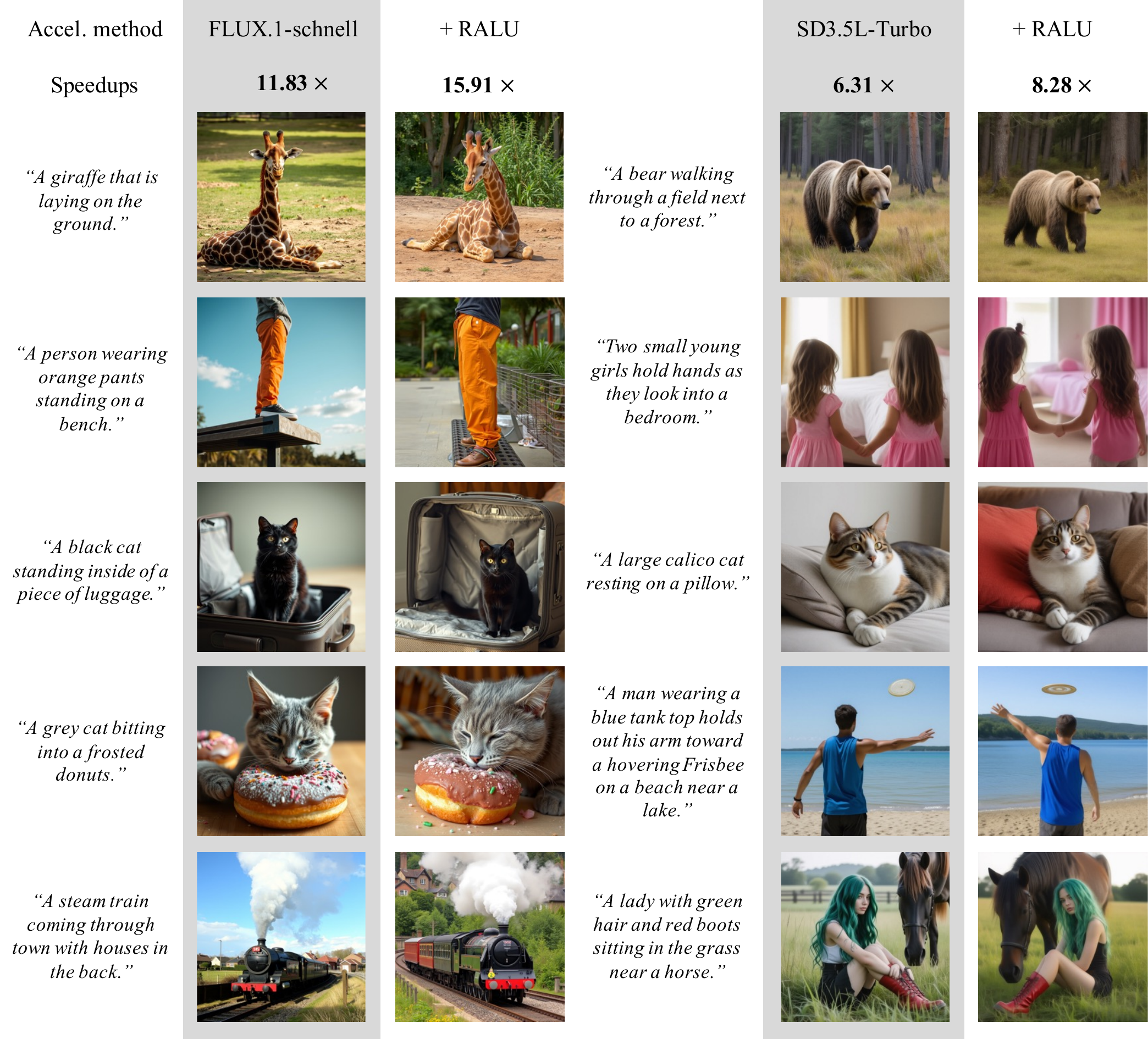}
    \caption{Qualitative comparison of timestep-distilled models integrated with RALU.}
    \label{fig:supp_distill_quali}
  \hfill
\end{figure*}

Fig.~\ref{fig:supp_distill_quali} presents qualitative comparisons obtained by integrating RALU with the timestep-distilled models FLUX.1-schnell and SD3.5L-Turbo. Both models operate with an NFE of 4 in combination with RALU, using $[N_1, N_2, N_3] = [1, 2, 4]$ for each. The results show that, even when achieving up to a 15.91$\times$ speedup through this integration, the image quality remains largely unaffected and no noticeable artifacts are introduced.

\subsection{Additional Qualitative Results}
\begin{figure*}[t]
  \centering
  \includegraphics[width=0.9\linewidth]{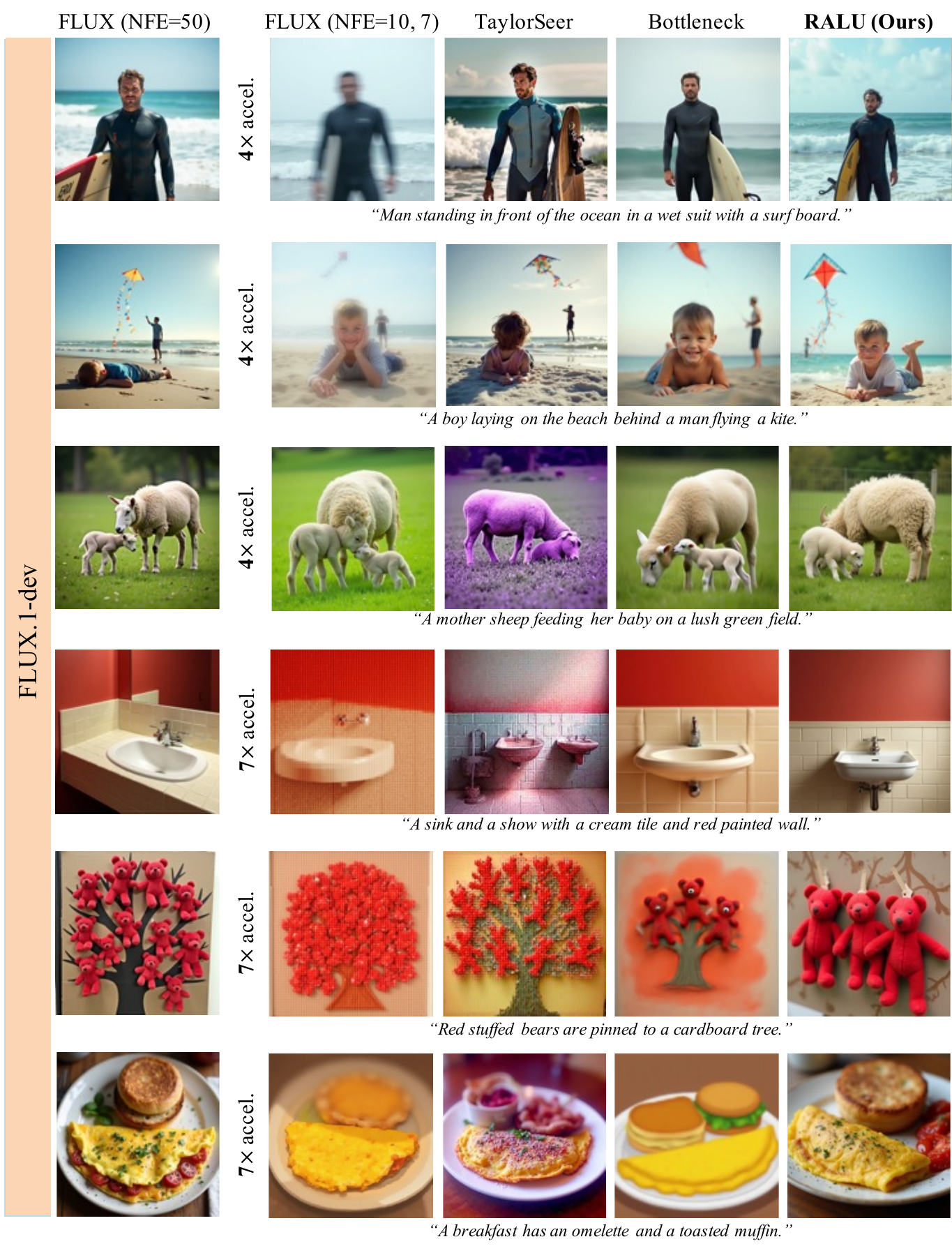}
    \caption{Qualitative comparison of images generated by baseline methods and RALU on FLUX. Best viewed in zoom.}
    \label{fig:quali_supp_flux}
  \hfill
\end{figure*}

\begin{figure*}[!t]
  \centering
  \includegraphics[width=0.86\linewidth]{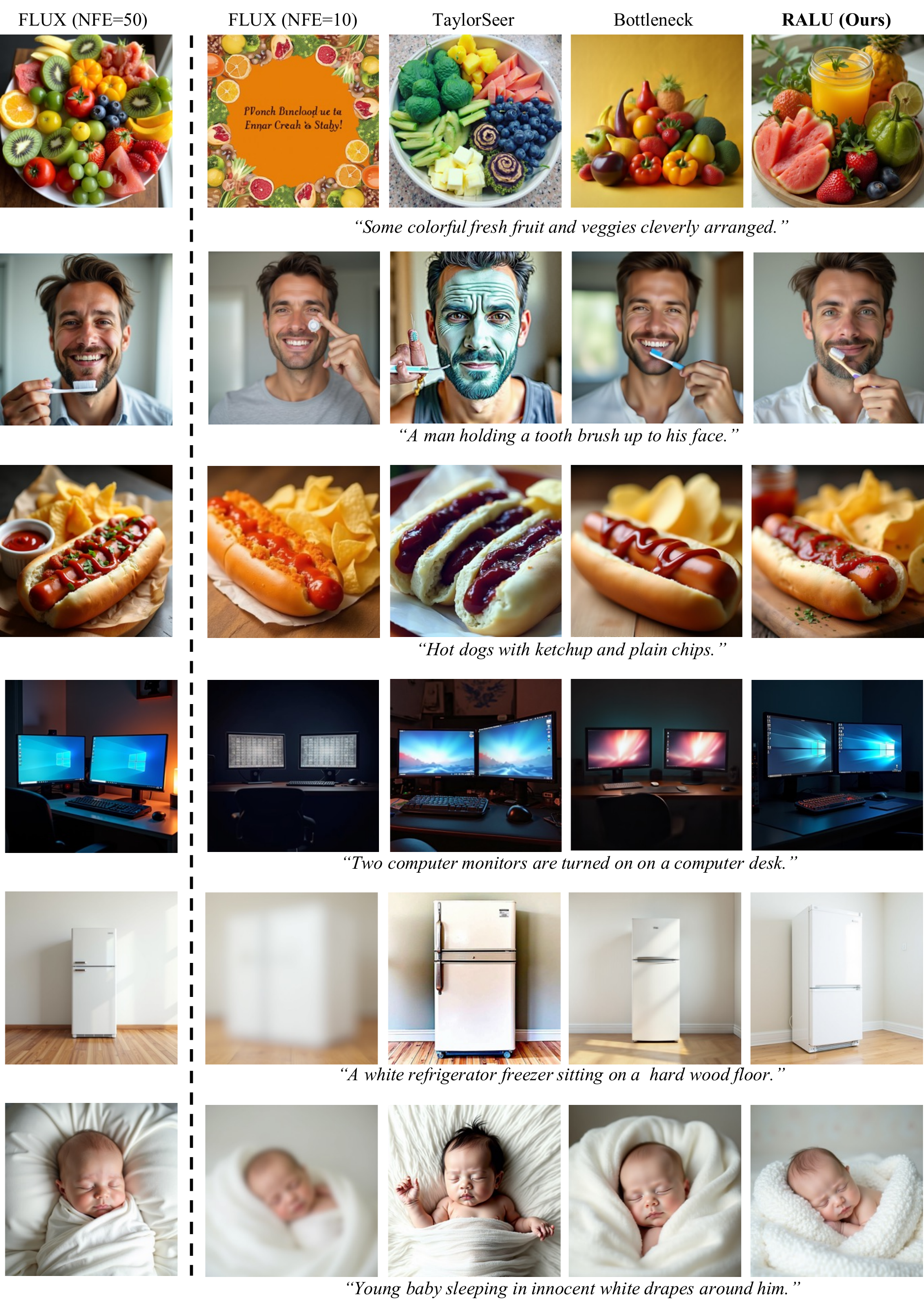}
    \caption{Qualitative comparison of images generated by baseline methods and RALU on FLUX for 5$\times$ speedups. Best viewed in zoom.}
    \label{fig:qualitative_flux_4}
  \hfill
\end{figure*}

\begin{figure*}[!t]
  \centering
  \includegraphics[width=0.86\linewidth]{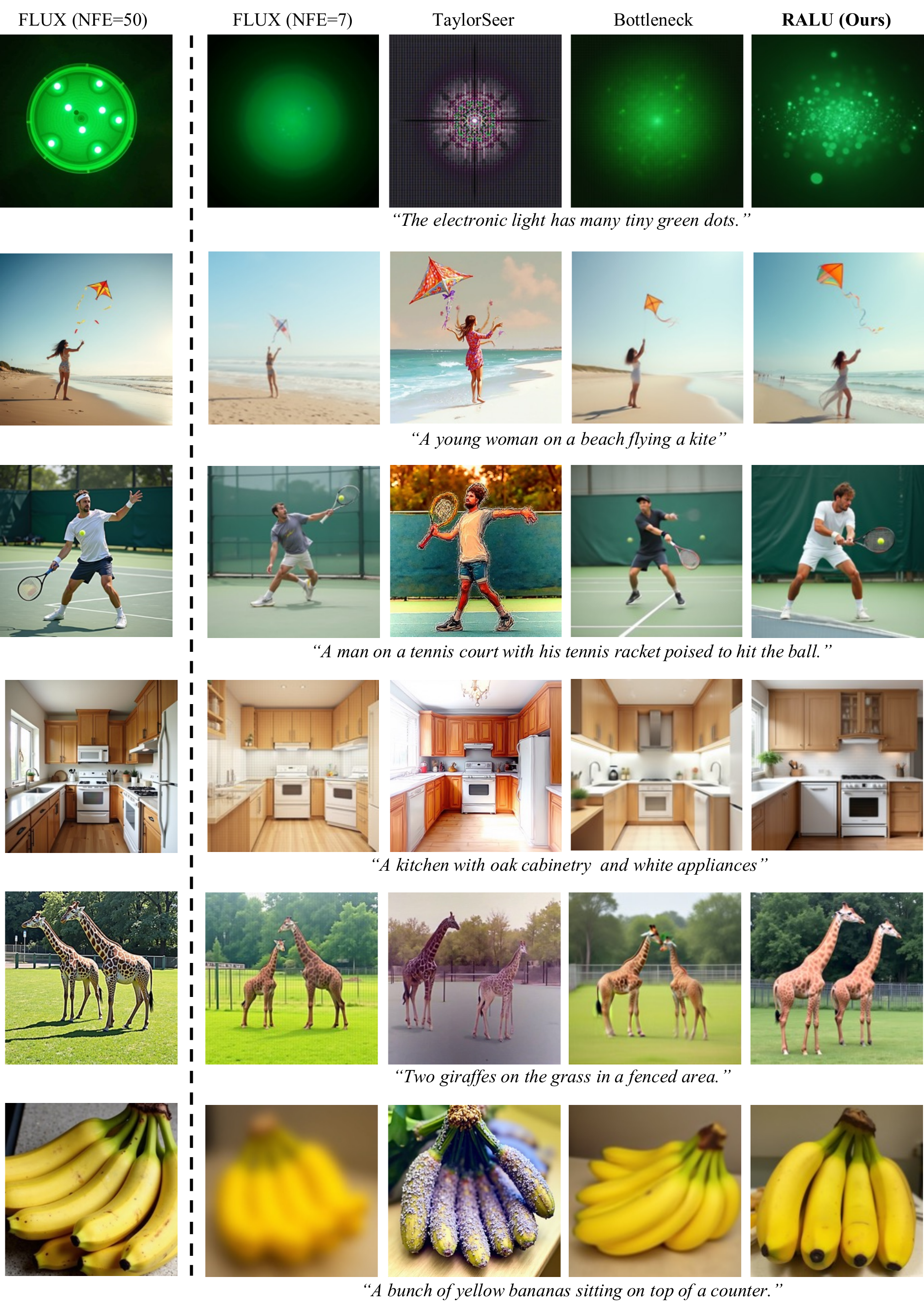}
    \caption{Qualitative comparison of images generated by baseline methods and RALU on FLUX for 7$\times$ speedups. Best viewed in zoom.}
    \label{fig:qualitative_flux_7}
  \hfill
\end{figure*}

\begin{figure*}[t]
  \centering
  \includegraphics[width=0.9\linewidth]{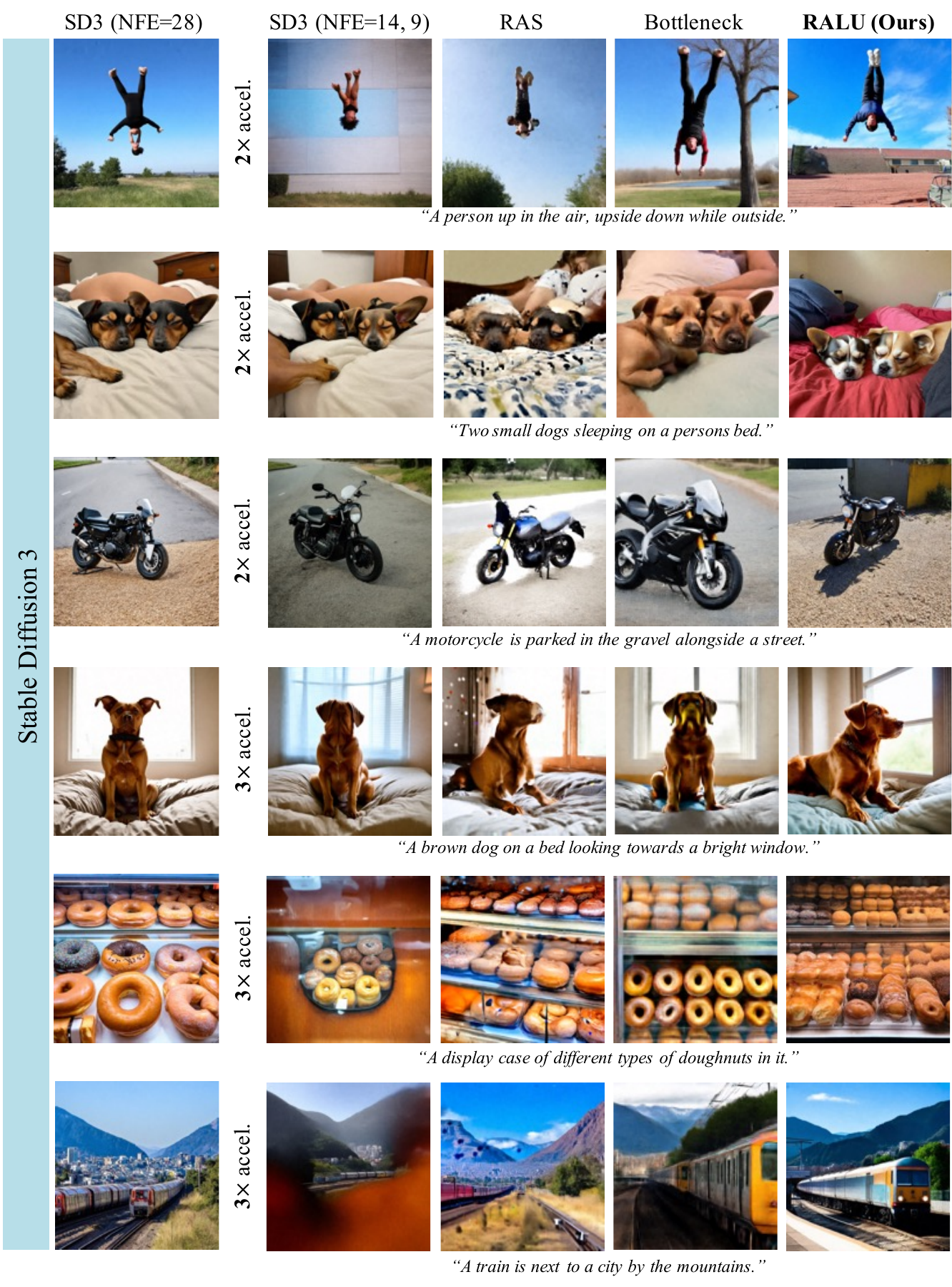}
    \caption{Qualitative comparison of images generated by baseline methods and RALU on SD3. Best viewed in zoom.}
    \label{fig:quali_supp_sd3}
  \hfill
\end{figure*}

\begin{figure*}[!t]
  \centering
  \includegraphics[width=0.86\linewidth]{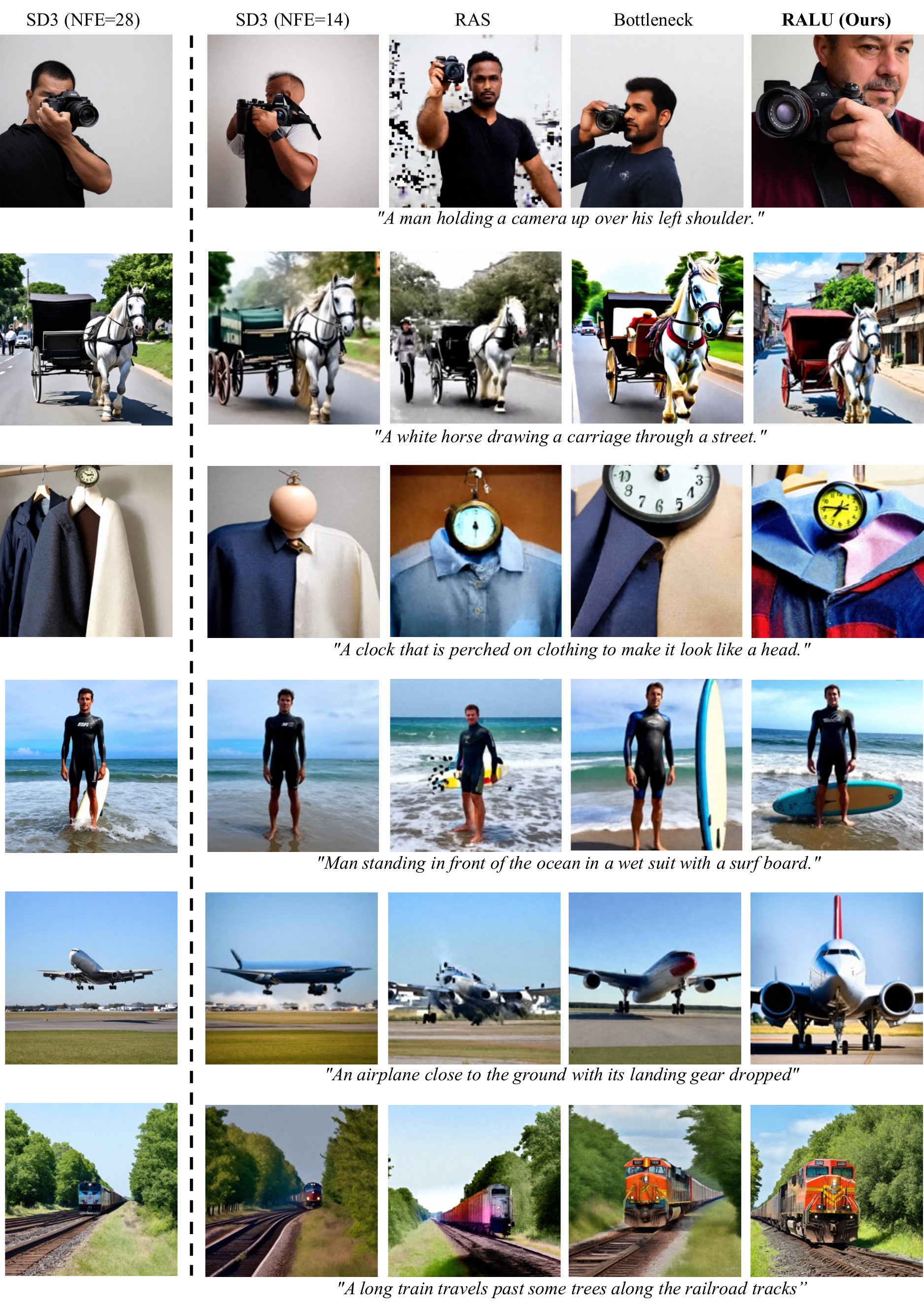}
    \caption{Qualitative comparison of images generated by baseline methods and RALU on SD3 for 2$\times$ speedups. Best viewed in zoom.}
    \label{fig:qualitative_sd3_2}
  \hfill
\end{figure*}

\begin{figure*}[!t]
  \centering
  \includegraphics[width=0.86\linewidth]{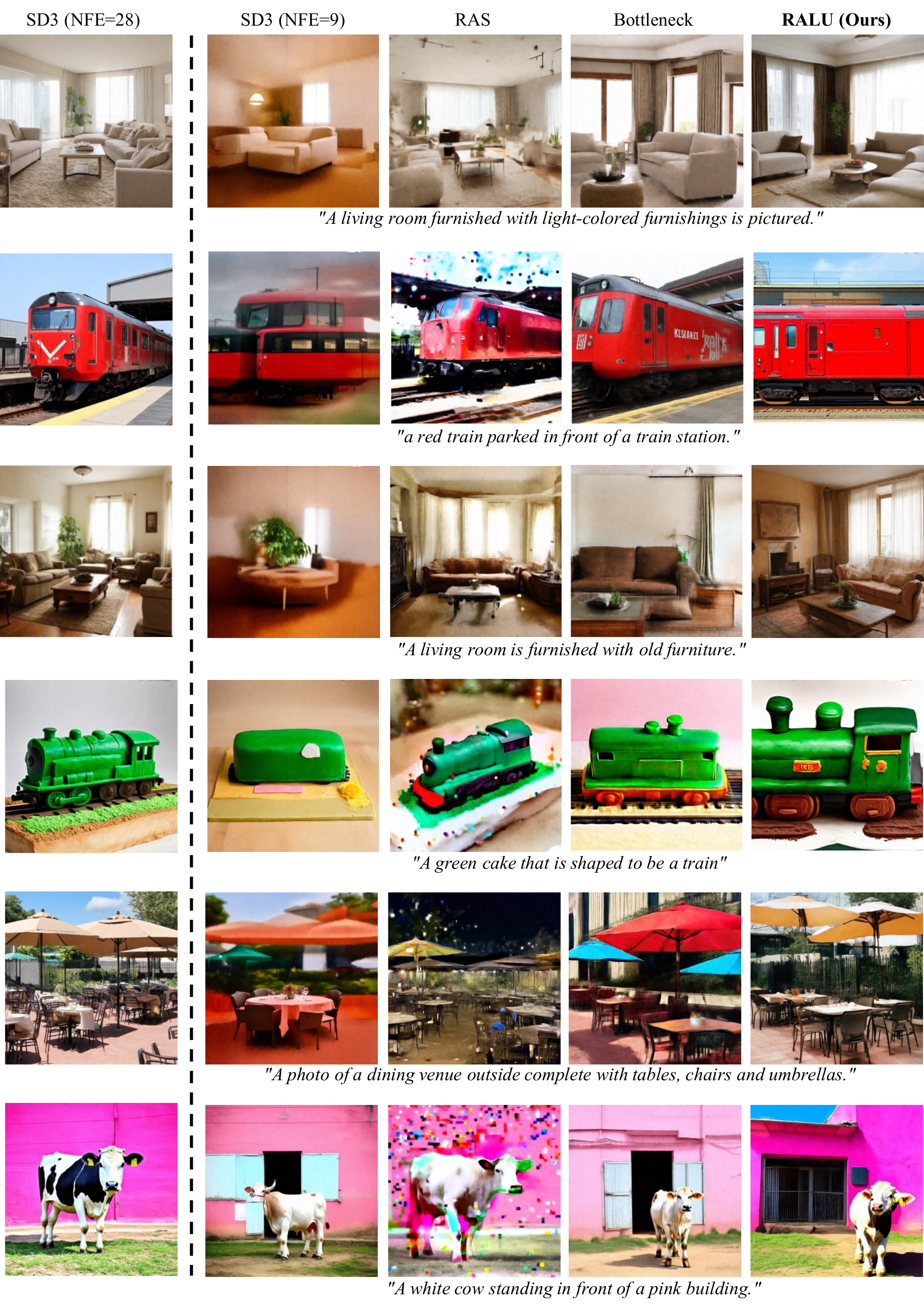}
    \caption{Qualitative comparison of images generated by baseline methods and RALU on SD3 for 3$\times$ speedups. Best viewed in zoom.}
    \label{fig:qualitative_sd3_3}
  \hfill
\end{figure*}

Additional qualitative results are presented in Fig.~\ref{fig:quali_supp_flux}-~\ref{fig:qualitative_sd3_3} following the technical appendices. All experimental configurations are provided in the main paper and its appendices.
We prepared additional qualitative results following Fig.~6.

\subsection{Uncurated Qualitative Results}
\begin{figure*}[!b]
  \centering  \includegraphics[width=0.88\linewidth]{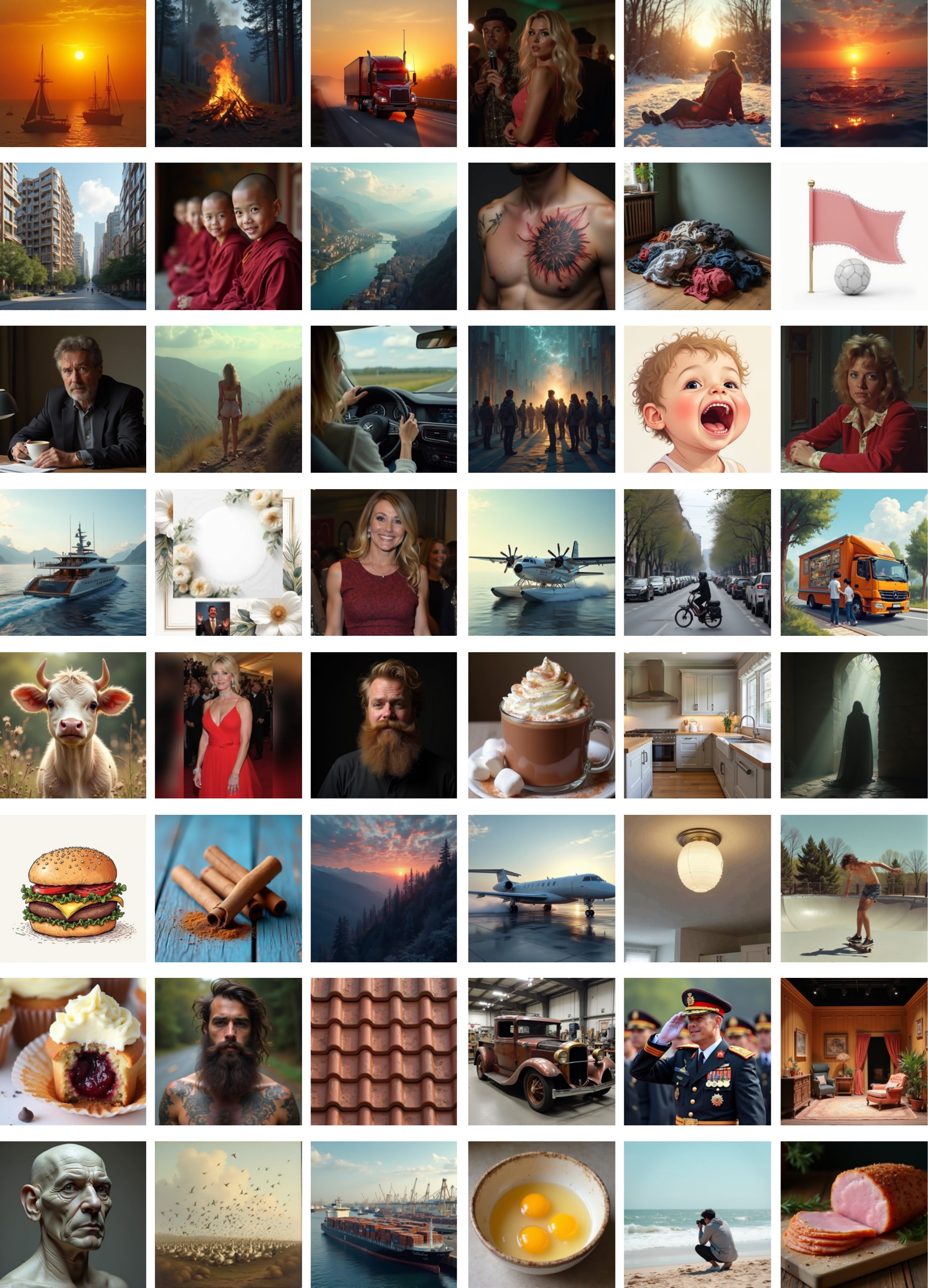}
    \caption{48 uncurated images generated by RALU on FLUX, 5$\times$ speedup.}
    \label{fig:uncurated_4x_1}
  \hfill
\end{figure*}

\begin{figure*}[t]
  \centering
  \includegraphics[width=0.88\linewidth]{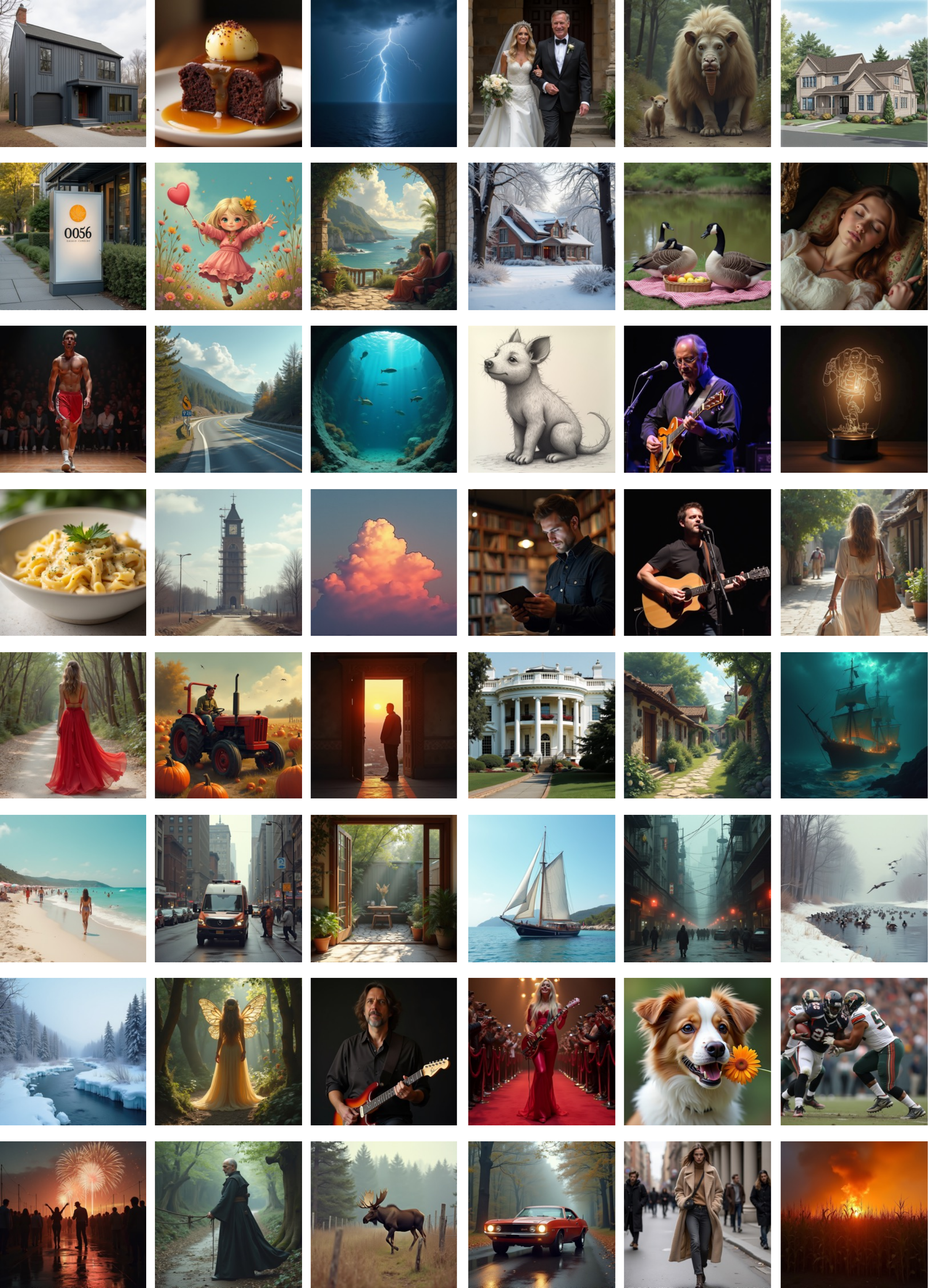}
    \caption{48 uncurated images generated by RALU on FLUX, 5$\times$ speedup.}
    \label{fig:uncurated_4x_2}
  \hfill
\end{figure*}

\begin{figure*}[t]
  \centering
  \includegraphics[width=0.88\linewidth]{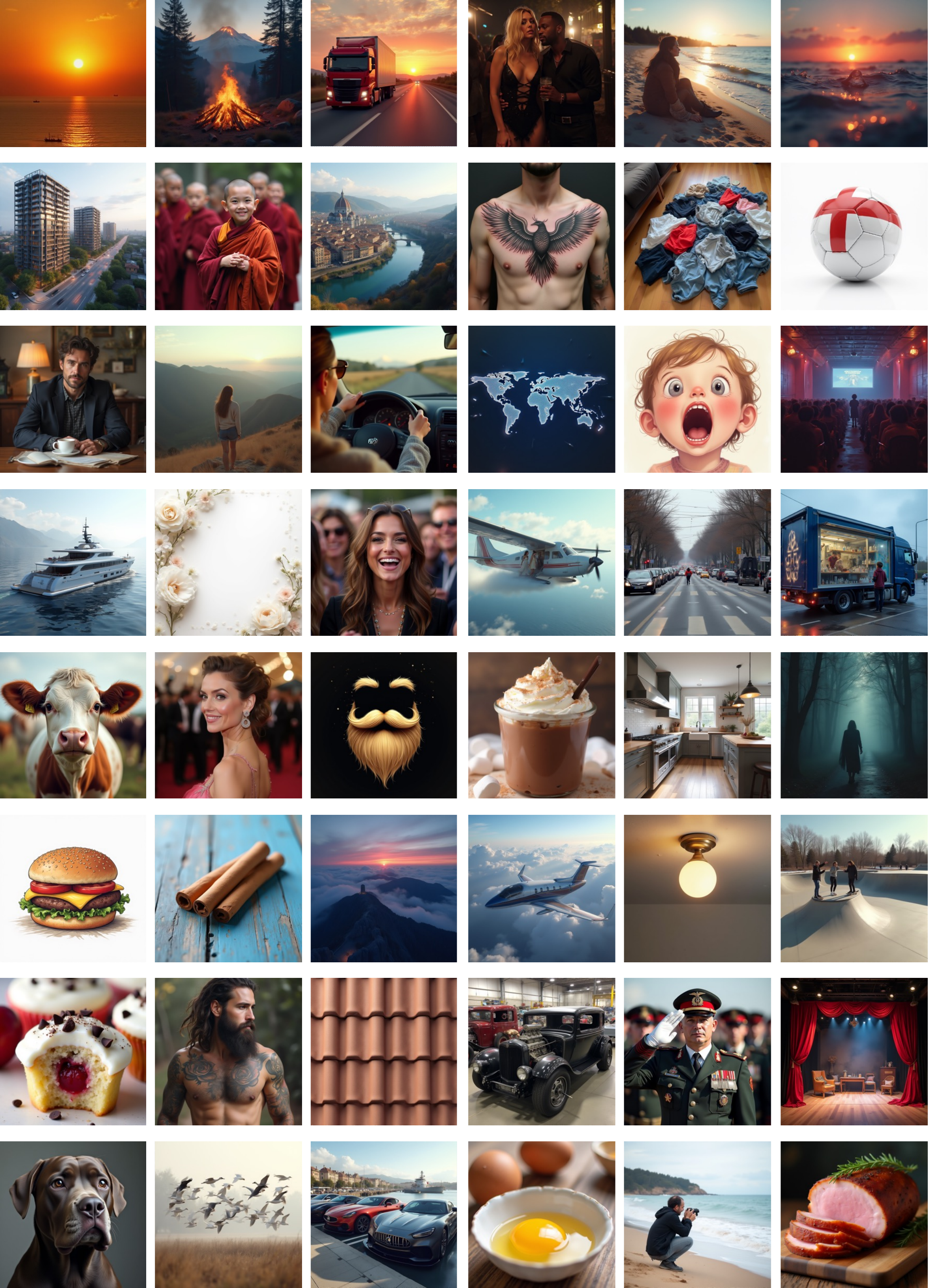}
    \caption{48 uncurated images generated by RALU on FLUX, 7$\times$ speedup.}
    \label{fig:uncurated_7x_1}
  \hfill
\end{figure*}

\begin{figure*}[t]
  \centering
  \includegraphics[width=0.88\linewidth]{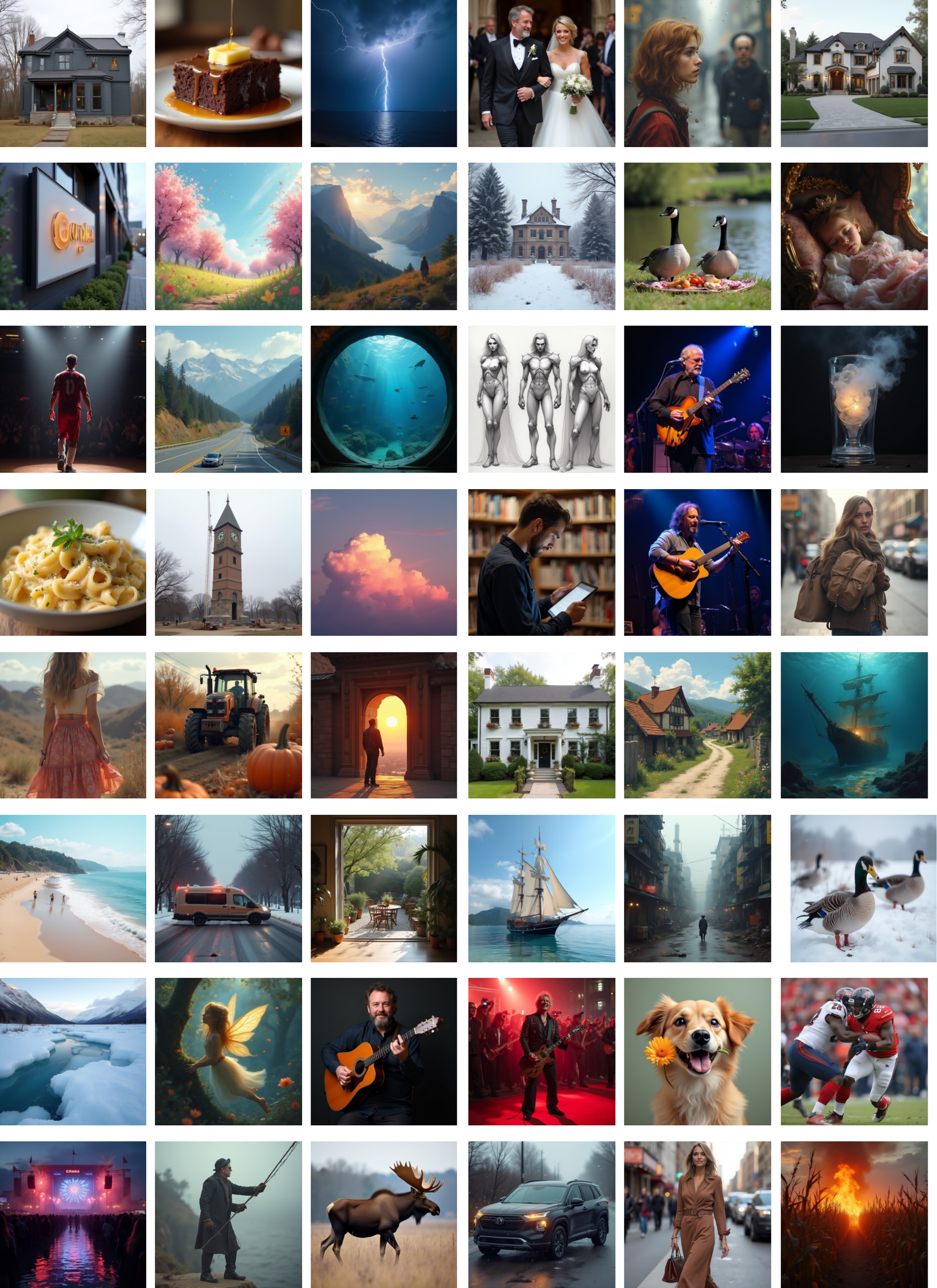}
    \caption{48 uncurated images generated by RALU on FLUX, 7$\times$ speedup.}
    \label{fig:uncurated_7x_2}
  \hfill
\end{figure*}

To demonstrate that our model consistently maintains high generation quality without cherry-picking, we present uncurated qualitative results under 5$\times$ and 7$\times$ speedup on FLUX. We randomly sampled 96 prompts from the CC12M~\cite{changpinyo2021conceptual} dataset and generated corresponding images. The results are shown in Fig.~\ref{fig:uncurated_4x_1}-~\ref{fig:uncurated_7x_2}.

\section{Limitations}
\label{Limitations}
While region-adaptive early upsampling is broadly applicable to diffusion transformers (DiTs), the Noise-Timestep Matching (NT-Matching) component is tailored specifically for flow-matching–based architectures. Moreover, our method is evaluated only on DiTs; its applicability to other generative backbones, such as diffusion U-Net, remains unverified. Consequently, the generality of NT-Matching beyond flow-matching DiTs has yet to be established.

\section{Broader Impact}
\label{broader_impact}
RALU enables faster and more resource-efficient generation of high-quality images using diffusion transformers, which has the potential to make such models more accessible for real-world or on-device applications. This could democratize creative tools for broader user groups while reducing environmental costs associated with large-scale inference. However, this efficiency gain may also facilitate misuse, such as faster generation of harmful or misleading content. Additionally, the selective focus on visually salient regions (e.g., edges) may implicitly encode or reinforce dataset biases, especially in underrepresented object structures. Care must be taken to evaluate fairness and misuse risks, and we encourage future work to explore responsible deployment strategies alongside technical improvements.

\clearpage

\clearpage

{
    \small
    \bibliographystyle{ieeenat_fullname}
    \bibliography{main}
}

\end{document}